%% file: acl-main.tex
\pdfoutput=1

\documentclass[11pt]{article}

\usepackage[preprint]{acl}
\usepackage{times}
\usepackage{latexsym}
\usepackage{xspace}

\usepackage[T1]{fontenc}

\usepackage[utf8]{inputenc}

\usepackage{microtype}
\usepackage{makecell}
\usepackage{inconsolata}
\usepackage{tcolorbox}

\usepackage{graphicx}
\usepackage{amsmath}
\usepackage{booktabs}
\usepackage{multicol}
\usepackage{multirow}
\usepackage{array}

\usepackage{subfigure}      %
\usepackage{caption}     %
\usepackage{subcaption}
\usepackage{hyperref}
\usepackage{textcomp}

\newcommand{\oursys}[1]{\textbf{PANDA}}

\newcommand{\agentopeh}{\(A_{\text{Ope.}^{\uparrow}}\)}
\newcommand{\agentopel}{\(A_{\text{Ope.}^{\downarrow}}\)}
\newcommand{\agentconh}{\(A_{\text{Con.}^{\uparrow}}\)}
\newcommand{\agentconl}{\(A_{\text{Con.}^{\downarrow}}\)}
\newcommand{\agentexth}{\(A_{\text{Ext.}^{\uparrow}}\)}
\newcommand{\agentextl}{\(A_{\text{Ext.}^{\downarrow}}\)}
\newcommand{\agentagrh}{\(A_{\text{Agr.}^{\uparrow}}\)}
\newcommand{\agentagrl}{\(A_{\text{Agr.}^{\downarrow}}\)}
\newcommand{\agentneuh}{\(A_{\text{Neu.}^{\uparrow}}\)}
\newcommand{\agentneul}{\(A_{\text{Neu.}^{\downarrow}}\)}
\newcommand{\agentpsyh}{\(A_{\text{Psy.}^{\uparrow}}\)}
\newcommand{\agentpsyl}{\(A_{\text{Psy.}^{\downarrow}}\)}
\newcommand{\agentmach}{\(A_{\text{Mac.}^{\uparrow}}\)}
\newcommand{\agentmacl}{\(A_{\text{Mac.}^{\downarrow}}\)}
\newcommand{\agentnarh}{\(A_{\text{Nar.}^{\uparrow}}\)}
\newcommand{\agentnarl}{\(A_{\text{Nar.}^{\downarrow}}\)}
\newcommand{\agentnp}{\(A_{\text{N.P}}\)}

\newcommand{\actionhigh}{\( a_{p\text{\textuparrow }} \)}
\newcommand{\actionlow}{\( a_{p\text{\textdownarrow }} \)}

\newcommand{\agenthigh}{\(A_{p^{\uparrow}}\)}
\newcommand{\agentlow}{\(A_{p^{\downarrow}}\)}

\newcommand{\huggingface}{\raisebox{-1.5pt}{\includegraphics[height=1.05em]{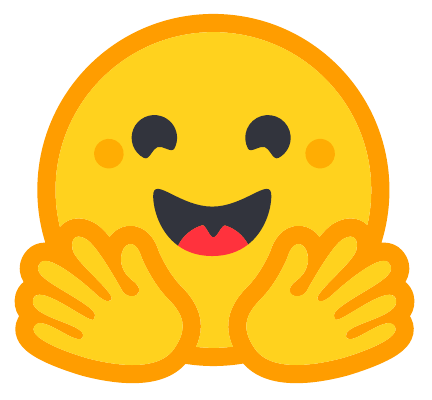}}\xspace}
\newcommand{\github}{\raisebox{-1.5pt}{\includegraphics[height=1.05em]{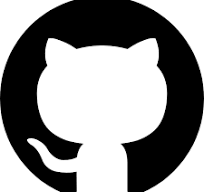}}\xspace}

\title{Persona Dynamics: Unveiling the Impact of Personality Traits on Agents in Text-Based Games}

\author{
\textbf{Seungwon Lim} \hspace{0.5cm}
\textbf{Seungbeen Lee} \hspace{0.5cm}
\textbf{Dongjun Min} \hspace{0.5cm}
\textbf{Youngjae Yu}\textsuperscript{\textdagger} \\[0.3cm]
{Yonsei University}
\\
  \small{
    \href{mailto:email@domain}{sngwon@yonsei.ac.kr}
  }
}

\begin{document}
\maketitle
\begin{abstract}
Artificial agents are increasingly central to complex interactions and decision-making tasks, yet aligning their behaviors with desired human values remains an open challenge. In this work, we investigate how human-like personality traits influence agent behavior and performance within text-based interactive environments. We introduce \oursys{}: \textbf{P}ersonality-\textbf{A}dapted \textbf{N}eural \textbf{D}ecision \textbf{A}gents, a novel method for projecting human personality traits onto agents to guide their behavior. To induce personality in a text-based game agent, (i) we train a personality classifier to identify what personality type the agent's actions exhibit, and (ii) we integrate the personality profiles directly into the agent’s policy-learning pipeline. By deploying agents embodying 16 distinct personality types across 25 text-based games and analyzing their trajectories, we demonstrate that an agent's action decisions can be guided toward specific personality profiles. Moreover, certain personality types, such as those characterized by higher levels of Openness, display marked advantages in performance. These findings underscore the promise of personality-adapted agents for fostering more aligned, effective, and human-centric decision-making in interactive environments.
\footnote{
  \textdagger denotes corresponding author. \\
  \github \textbf{Code}: \href{https://github.com/pull-ups/PANDA}{pull-ups/PANDA}\\
  \huggingface \textbf{Model}: \href{https://huggingface.co/mirlab/PersonalityClassifier}{mirlab/PersonalityClassifier}}

\end{abstract}

\section{Introduction}\label{sec:intro}

Text-based interactive environments, exemplified by text-based games, have long presented formidable challenges for AI~\cite{lin2024swiftsage, yao2020keep}. Unlike traditional games such as Atari, chess, and Go where the possible spaces for action and environment are predefined and effective actions can be learned based on statistics, playing text-based games requires a combination of complex skills related to natural language processing. These skills include understanding the environment and generating appropriate actions in response, both presented in the textual description.

Recent advances in Large Language Models (LLMs) have expanded their utility beyond traditional closed-set evaluations on fixed benchmarks~\cite{hendrycks2020measuring, srivastava2022beyond}, leading to growing interest in validating these models' capabilities in interactive environments~\cite{ahn2022can, yao2023react}. These scenarios require both environmental interaction and decision-making abilities. Text-based games—where a series of actions must be evaluated through interaction with the environment—serve as an excellent testbench for verifying these capabilities.

\begin{figure}[!t]
    \centering
    \includegraphics[width=\columnwidth]{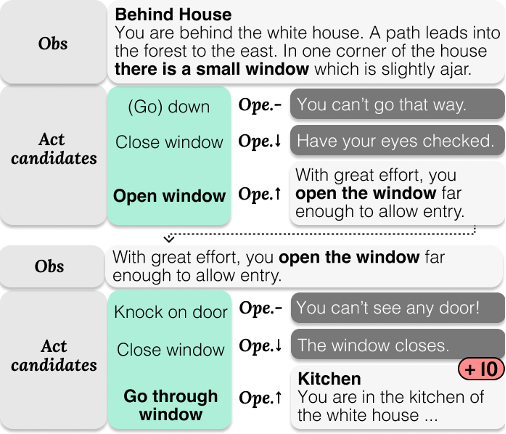}
    \caption{Excerpt from Jiminy Cricket benchmark (`Zork1'). The action of high openness (annotated by our classifier) leads the player to explore new areas (\texttt{Open window}) and progress (\texttt{Go through window}).}
    \label{fig:baselineoverall}
\end{figure}

Initial efforts in this domain concentrated on improving performance through Reinforcement Learning approaches~\cite{he2016deep, yao2020keep}. More recently, attention has turned to integrating human values into agent behavior. For example, ~\cite{hendrycks2021what, pan2023rewards} instill ethics and morals in agents, while ~\cite{ammanabrolu2022aligning} instills social norms. While these works focus on universal value systems, they have not yet explored the role of diverse intrinsic traits in guiding agent behavior.

\begin{figure*}[!t]
    \centering
    \includegraphics[width=\textwidth, height=\textheight, keepaspectratio]{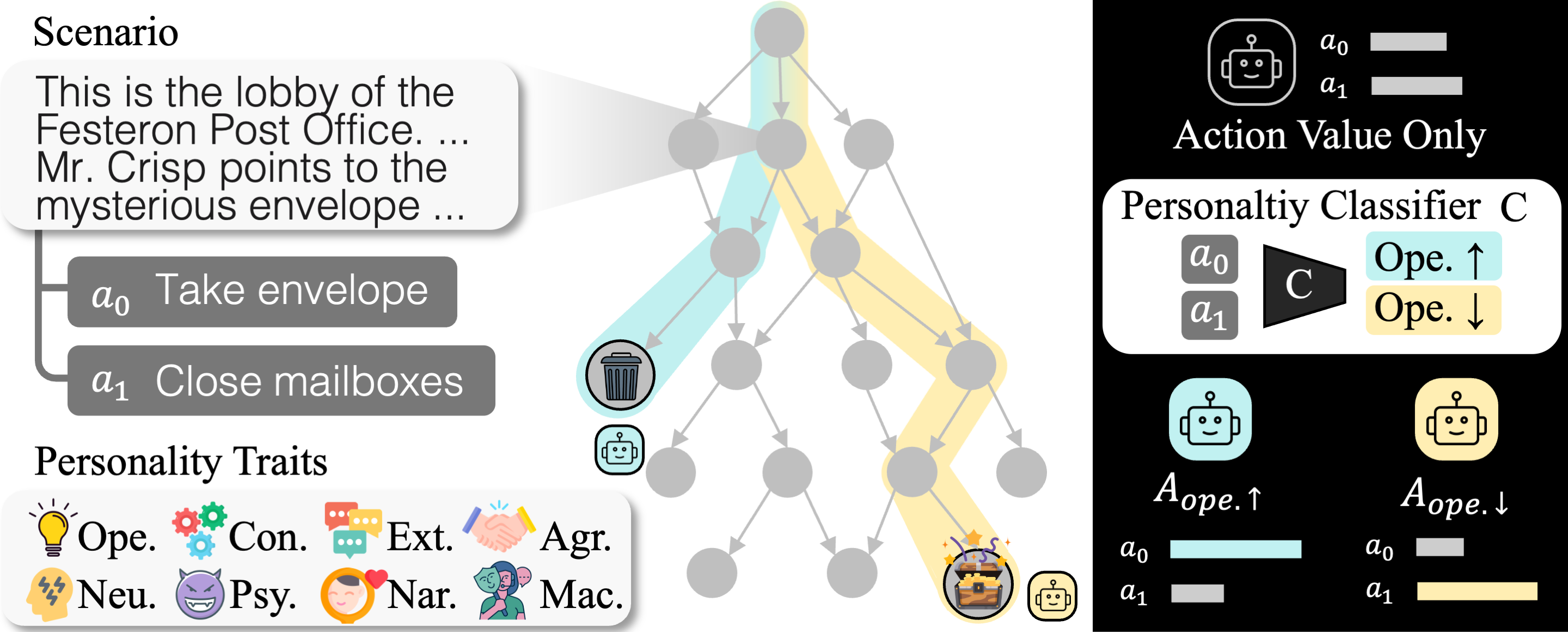}
    \caption{Mock-up of a game in the game of Jiminy Cricket benchmark. Within the game, each place, interactable objects, and situation is defined as a node, and the agent can visit different nodes by doing action, which is generating textual action in text-based game. When visiting specific nodes and perform specific interaction, the game score increases and the agent receives a reward.}
    \label{fig:panda_main}
\end{figure*}

In this work, we expand the notion of “values” to include a broader spectrum of internal characteristics—namely, \textbf{personality} traits.  Our approach, \oursys{}, encompasses eight distinct of personality dimensions, including both the Big Five factors~\cite{John1999TheBF} and the Dark Triad~\cite{jones2014introducing}, This holistic viewpoint allows us to consider not only ethical or moral qualities but also other intrinsic traits that influence how agents perceive and interact with their environments.

We employ the Jiminy Cricket benchmark~\cite{hendrycks2021what}, a suite of 25 complex text-based adventure games spanning over 1,800 locations and nearly 5,000 interactable objects. This rich environment provides ample scope to observe how different personality types affect exploration and problem-solving. In particular, we find that agents exhibiting high Openness—characterized by curiosity, adventurousness, and a propensity for novel experiences~\cite{dumblekar2024openness, bateman2016aesthetic}—consistently engage in more extensive exploration, undertake more interactions, and achieve higher game scores.

By incorporating the personality dimension into the evaluation of artificial agents in interactive environments, our research aims to provide new perspectives on how personality traits can be leveraged to affect an agent's action decision and improve performance.
This work contributes to a broader understanding of AI behavior in complex narrative settings, advancing the development of agents that not only act morally and socially acceptably, but also exhibit specific personality traits.

\section{Personality Guidance in Textgame}
\subsection{Text-based Game}\label{sec:textgame}

A text-based game can be formally specified as a partially observable Markov decision process (POMDP) $(\mathcal{S}, \mathcal{T}, \mathcal{A}, \mathcal{O}, \mathcal{R})$. For the latent current game state $\mathcal{S}$, which contains internal information such as the agent's location and stats, the agent receives information that is currently observable to it in the form of a text paragraph observation $\mathcal{O}$. The agent then performs an action $\mathcal{A}$ in text form, which changes $\mathcal{S}$ to $\mathcal{S}'$ according to the transition function $\mathcal{T}: \mathcal{S} \times \mathcal{A} \rightarrow \mathcal{S}'$. If a predefined condition-satisfying action is performed in a specific state $\mathcal{S}^\ast$, a reward is given to the agent, and the game score is calculated with the sum of rewards within a single trajectory.

\subsection{Environment}\label{sec:environment}
To explore the action patterns of agents with different personalities and traits in adventurous text-based games and to examine the differences between them, we utilize The Jiminy Cricket~\cite{hendrycks2021what} benchmark. It is composed of 25 text-based games based on the interpreter of Jericho~\cite{hausknecht2020interactive}, which is designed for studying and evaluating agent performance in an adventurous environment. In Jiminy Cricket's games, actions are defined as free-form text where only admissible actions determined by internally defined parsing rules (PDDL) induce state transitions.

\subsection{Agent Implementation}\label{sec:agent_implementation}
In this paper, the overall agent architecture in both benchmarks is implemented upon the Deep Reinforcement Relevance Network (DRRN)~\cite{he2016deep}, which has been commonly adopted as the primary framework for text-based game learning~\cite{ammanabrolu2022aligning, hendrycks2021what, yao-etal-2020-keep}. 

In DRRN, the neural network is trained to predict Q-value \( Q(s_t, a_t) \), the action-value function for actions in game states at time step $t$, utilizing deep Q-learning~\cite{watkins1992q}. The policy $\pi(a_t\mid c_t)$ is configured to select the action that maximizes this value.

\begin{figure}[!t]
    \centering
    \includegraphics[width=\columnwidth]{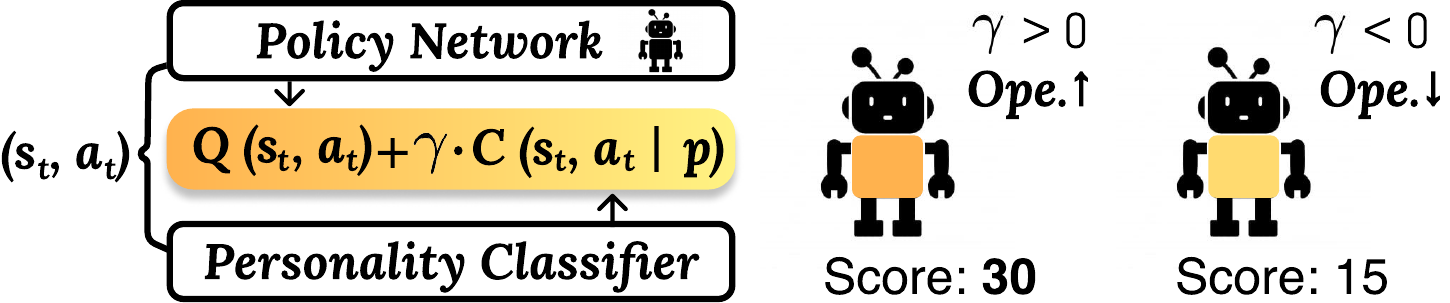}
    \caption{\oursys{} Framework. At each steps' state \(s_t\), agents are guided by both the Q-values from the policy network and the valence values derived from the personality classifier.}
    \label{fig:panda_framework}
\end{figure}

To guide the agents to perform an action that aligns with personality, we use the result from a personality classifier (\S\ref{sec:personality classifier}) to guide the agent's behavior, as illustrated in Figure~\ref{fig:panda_framework}.  Specifically, adjusted Q-value \( Q'(s_t, a_t) \) is calculated by equation (\ref{eq:policy_shaping}):
\begin{equation}
Q'(s_t, a_t^i) = Q(s_t, a_t^i)  +  \gamma * C(s_t, a_t^i \mid p) \\
\label{eq:policy_shaping}
\end{equation}

where \( Q(s_t, a_t^i) \) is the action value of i-th action among action candidates, optimized during training. \( C(s_t, a_t^i \mid p ) \in \{-1, 0, 1\} \) represents the output of personality classifier. Given a pair of situation \(s_t\) and action \(a_t^i\), \(-1\) denotes \textit{Low} valence, \(0\) denotes \textit{Neutral} valence, and \(1\) denotes \textit{High} valence regarding personality type $p$ to evaluate. The sign of \(\gamma\) determines the direction of alignment. When \(\gamma\) \(>0\), it increases the Q-value of behaviors that match the personality trait, and vice versa. The absolute value determines the strength of the intended alignment. The agent's action selection \(a_t\) at state \(s_t\) is determined by (\ref{eq:action_prob}):
\begin{equation}
\pi(a_t = a_t^i|s_t) = \frac{\exp(Q(s_t, a_t^i))}{\sum_{j=1}^{|\mathcal{A}_t|}\exp(Q(s_t, a_t^j))}
\label{eq:action_prob}
\end{equation}
where \(|\mathcal{A}_t|\) denotes total number of action candidates. Qualitative examples of actions selected by agents with different personality traits are in Table~\ref{tab:personality_qual}.
\paragraph{Notation}

In this paper, we denote agents under the guidance of specific personality \(p\) as \agenthigh \ when \(\gamma\) \(>0\) and \agentlow\ when \(\gamma\) \(<0\) in equation (\ref{eq:policy_shaping}), reflecting the valence of personality to guide. For example, an agent with high openness and low openness is denoted by \agentopeh{} and \agentopel{}, respectively.
Similarly, for a specific personality \(p\), an action where the classifier predicted as \textit{High} is denoted by \actionhigh, and \actionlow when \textit{Low}.

\section{Personality Classifier}\label{sec:personality classifier}

To guide an agent’s behavior according to a specified personality profile, we introduce a \textbf{personality classifier guidance} mechanism. This approach enables the agent to incorporate personality-related considerations into its learning process, ensuring that its actions align with the desired personality traits.

To train this personality classifier, we first construct a large-scale dataset of 120,000 personality-labeled examples using GPT-4~\cite{achiam2023gpt} (Sec.~\ref{sec:datamake}). We then fine-tune a Flan-T5-XL~\cite{chung2024scaling}, which has 3 billion parameters and provides efficient inference (Sec~\ref{sec:classifier}). The resulting classifier achieves a high degree of accuracy (98.59\% as shown in Table~\ref{tab:classifier_performance}).

\subsection{Dataset Construction}\label{sec:datamake}
\paragraph{Starting From Validated Personality Descriptions.}

We begin by employing the widely-adopted Big Five~\cite{mccrae1987validation} and Dark Triad~\cite{paulhus2014toward}, to characterize game agents by eight distinct personality traits (see Table \ref{tab:abbreviations} for abbreviations). Following the methodology of \citet{lee2024llms}, we use items of validated questionnaire, BFI~\cite{John1999TheBF} and SD-3~\cite{jones2014introducing}, as the foundation for dataset expansion.

Although these sentences collectively address various facets of each trait, there is a noticeable imbalance in their distribution, and approximately 70\% of the items describe individuals exhibiting a high valence toward a given trait. To achieve balance, we systematically paraphrase these initial sentences to create 10 instances per trait (5 representing high valence and 5 representing low valence).
Through this process, we obtain a total of 80 descriptions across 8 personality types Full examples are in~\ref{tab:paraphrase1} and~\ref{tab:paraphrase2}.

\input{tables/personality_abbrv}

\paragraph{Augmentation with Situational Seeds.}

To make a single statement (e.g.`I easily make new friends.') into a detailed description (e.g. `I don't worry about making new friends when moving schools'), we use GPT-4 to generate 300 diverse, common situations. These are divided into 30 subsets (e.g. Home and Family), each containing 10 scenarios (e.g. Kitchen, Garden). This approach, despite known biases in GPT-4~\cite{gupta2023bias}, helps data augmentation with diversity with minimal duplication~\cite{west2021symbolic}. Using 80 personality-describing sentences across 8 traits and the 300 situational seeds, we generate 5 sentences for each seed combination ($80\times300\times5=120,000$). The intermediate results of the dataset creation process and examples of generated samples are in Table ~\ref{tab:dataset_example1} and~\ref{tab:dataset_example2}.

\subsection{Training and Performance}\label{sec:classifier}
We trained Flan-T5~\cite{chung2024scaling} of various sizes on the dataset created in \S3.1, and examined performance on personality classification using a diverse dataset which provided statements with annotated personality traits. Since BFI and SD-3 were used in dataset construction, we used IPIP~\cite{goldberg1999broad}, another personality questionnaire, and Essays dataset~\cite{pennebaker1999linguistic}, containing author personality annotations across various types of writings as out-of-domain evaluation sets.

For the task of predicting whether a personality trait's valence is high, low, or neutral when given a statement and personality type, Table~\ref{tab:classifier_performance} represents that Flan-T5 model series shows a robust classification capacity when trained with our personality data. Our method incorporates a classifier filtering approach that builds on the foundation of previous work using Transformer-based encoder-decoder models, specifically the T5 model \citep{raffel2023exploring}. While the prior work~\cite{ammanabrolu2022aligning} utilized Delphi~\cite{jiang2022machines}, a model trained on a diverse set of commonsense ethics datasets to provide value priors, our approach differentiates itself by focusing on personality-driven classification.

\input{tables/classifier_performance}

\input{tables/jericho_results}

\section{Results on Adventure Game}\label{sec:jericho_analysis}

We used the Jiminy Cricket benchmark~\cite{hendrycks2021what} to explore the action patterns of agents with different personality traits in adventurous text-based games and to examine the differences between them(\S\ref{sec:jmn-result}). We used a personality classifier(\S\ref{sec:personality classifier}) to impose personality constraints on the agent's decision-making(\S\ref{sec:agent_implementation}).

\subsection{Agent Implementation}\label{sec:agent_implementation}

\paragraph{Action Candidate Generator}
Since games in Jiminy Cricket benchmark require the user to input free-form actions but only a limited number of them are valid, It is unsuitable to use an off-the-shelf LLM without any adaptation to the game environment. So we use an action candidate generator~\cite{ammanabrolu2022aligning} to generate a set of state-appropriate actions that are likely to be valid within the game.

\subsection{Results}\label{sec:jmn-result}
Table~\ref{tab:jericho_score} presents the game results for 15 games from the Jiminy Cricket benchmark. Each scores are the averages of the last 50 episodes' scores with three different random seeds. To identify advantageous personality traits across diverse text adventure games $g$, we established three criteria: \\

\noindent\textbf{1. Counting (Cnt.)}
$$\sum_{g} \text{if }[s(v,p) > s(\text{‐}) \text{ and } s(\bar{v},p) < s(\text{‐})]$$
\textbf{2. Average Score (Avg.)}
$$\frac{1}{g}\sum_{g} s(v,p) < \frac{1}{g}\sum_{g} s(\text{‐}) < \frac{1}{g}\sum_{g} s(\bar{v},p)$$
\textbf{3. Difference (Diff.)}
$$s(v,p) - s(\bar{v},p)$$

where $s(\cdot)$ denotes the score from agent injected with a given personality trait ($s(\text{‐})$ represents their score from an agent without any personality traits), $p \in \{\textit{Agr, Con, Ext, Agr, Neu, Psy, Mac, Nar}\}$, $v \in \{\textit{high, low}\}$ and $\bar{v}$ denotes the complementary value of $v$.

Based on these criteria, we propose that \textbf{High Openness} leads to successful performance in text adventure games. Openness is characterized by creativity, curiosity, and a willingness to explore new ideas and experiences~\cite{mccrae1987creativity, dumblekar2024openness, bateman2016aesthetic}, and it can be particularly beneficial in text adventure games. 

\subsection{Statistical analysis}\label{sec:jmn-stat}
To examine whether openness increases game agents' performance and whether this effect is consistently applied across different games, we conducted various statistical analyses. Due to the non-parametric nature of game scores, we performed \textit{Wilcoxon signed-rank} and \textit{Friedman} test. For the \textit{Wilcoxon signed-rank}, we analyzed all possible pairs: (\agenthigh, \agentnp), (\agentnp, \agentlow), and (\agenthigh, \agentlow). For the \textit{Friedman} test, we analyzed (\agenthigh, \agentnp, \agentlow) collectively.

Table~\ref{tab:jericho_stat} shows that that openness demonstrates superior performance in both statistical metrics, with notably higher statistical values and significance levels.

\input{tables/jericho_stat_2}

\input{tables/slam}

\begin{figure}[t]
    \centering
    \includegraphics[width=\columnwidth]{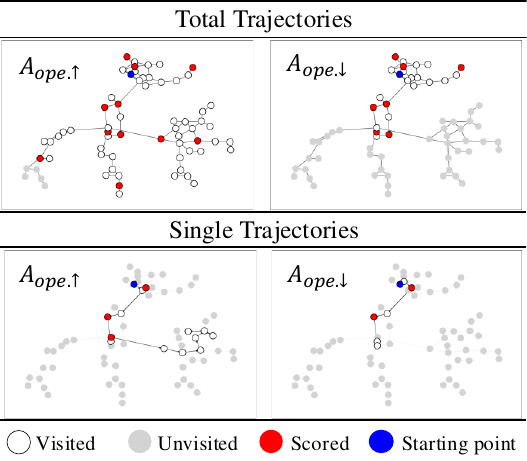}
    \caption{Trajectory Comparison between \agentopeh and \agentopel. For Total Trajectories, it shows all places visited by 8 multi-agents during 15,000 steps of training. Single Trajectory represents one example of these trajectories.}
    \label{fig:trajectory}
\end{figure}

\section{The Impact of Openness on Game Agents}\label{sec:jmn-analysis}
To achieve high performance in text adventure games provided by the Jiminy Cricket benchmark, it is essential to: \textbf{frequently visit reward-earning places}, and \textbf{perform reward-earning actions} at those locations. To analyze the positive impact of openness in text adventure games, we confirmed in \S\ref{sec:traj-analysis} and \S5.2 that agents with high openness traits excel in both aspects compared to other agents.

\subsection{Trajectory Analysis}\label{sec:traj-analysis}

In Table \ref{tab:slam}, we analyzed each agent's trajectory by categorizing locations into common (\textit{Com.}) and uncommon (\textit{Unc.}) places. From the starting point, locations with distances less than the specified depth were classified as \textit{Com.}, while the remaining locations were classified as \textit{Unc.} (See Appendix~\ref{sec:place_classification} for detailed difference). We analyzed both the visit counts and the number of steps required to reach these locations. 
Results show that \agentopeh visited the most spaces, which opens up possibilities for achieving high scores in the Zork1 game consisting of 110 locations. Additionally, in terms of Average steps, it showed the second shortest path after \agentexth, indicating that as a result of extensive exploration, it optimized travel paths to each location compared to the routes of other agents.

Figure~\ref{fig:trajectory} shows visual representations for each agent, with \agentopeh\ and \agentopel\ as representative examples. As shown in the \textit{Visit Count} of \agentopeh\ and \agentopel\ in Table~\ref{tab:slam}, both agents visited places near the starting point (\textit{Com.}) during their respective training periods (8.96 and 8.02). However, while \agentopel\ rarely reaches places far from the starting point (\textit{Unc.}), \agentopeh's trajectory branches out in multiple directions. The visual example of all agent types can be found in Figure~\ref{fig:traj_detail_1} to~\ref{fig:traj_detail_4}.

\subsection{Actions of Agent}
\subsubsection{Reward-earning Actions}
Even though an agent explores broad and diverse spaces, it must actually perform reward-earning actions to score points. To conduct a breakdown analysis of each agent's performance, we analyzed the reward-earning actions. 

Table~\ref{tab:qvalue} suggests that \agentopeh\ assigned higher values to reward-earning actions compared to other agents on average, and consequently performed more reward-earning actions during episodes, contributing to higher performance. Additionally, although \agentneuh\ showed the second-highest visits after \agentopeh in Table~\ref{tab:slam},  we can observe that it did not progress to performing many reward-earning actions.

\input{tables/qvalue}

\subsubsection{Alignment with given personality}
To verify whether agents assigned specific personality traits actually exhibited the intended behavioral patterns, we analyzed the distribution of actions by personality type that each game agent performed during training.

To analyze behavioral patterns induced by personality guidance, we normalized the number of actions performed by each agent (\agenthigh\ and \agentlow) using the actions of \agentnp\ as the baseline. 

Table~\ref{tab:action_ratio} demonstrates that all agents, except for \agentpsyh, exhibit behavior patterns that align with personality guidance. (positive for \agenthigh\ and negative for \agentlow.) However, this tendency diminishes as training progresses (from \textit{Init50} to \textit{Fin50}), suggesting that the personality guidance regulation, which was dominant in the early stages of training, becomes less strict as the policy network is optimized. However, in the case of \agentopeh, \agentnarh, \agentneul\ and \agentnarl, they learned to perform actions more aligned with their assigned personality during training (increased ratio for \agenthigh and decreased ratio for \agentlow).

\input{tables/action_ratio}

\subsection{Walkthrough Analysis}\label{sec:walkthrough_analysis}
Jiminy Cricket benchmark offers walkthroughs, which provide step-by-step guidance for optimal decision-making in each game scenario. Using GPT-4, we 
 analyzed the personality traits reflected in the actions composing the walkthroughs for all 25 games, by predicting which of the 16 personality types most closely matches the personality tendencies exhibited by the actions. Table~\ref{tab:walkthrough} shows that a high level of openness is most commonly required for agents to achieve successful outcomes across the games, highlighting the effect of personality guidance toward high openness. 

\input{tables/personality_walkthrough}

\subsection{Learning Curve Analysis}\label{sec:jmn-result-opn}

\begin{figure*}[htp!]
    \centering
    \includegraphics[width=0.9\textwidth]{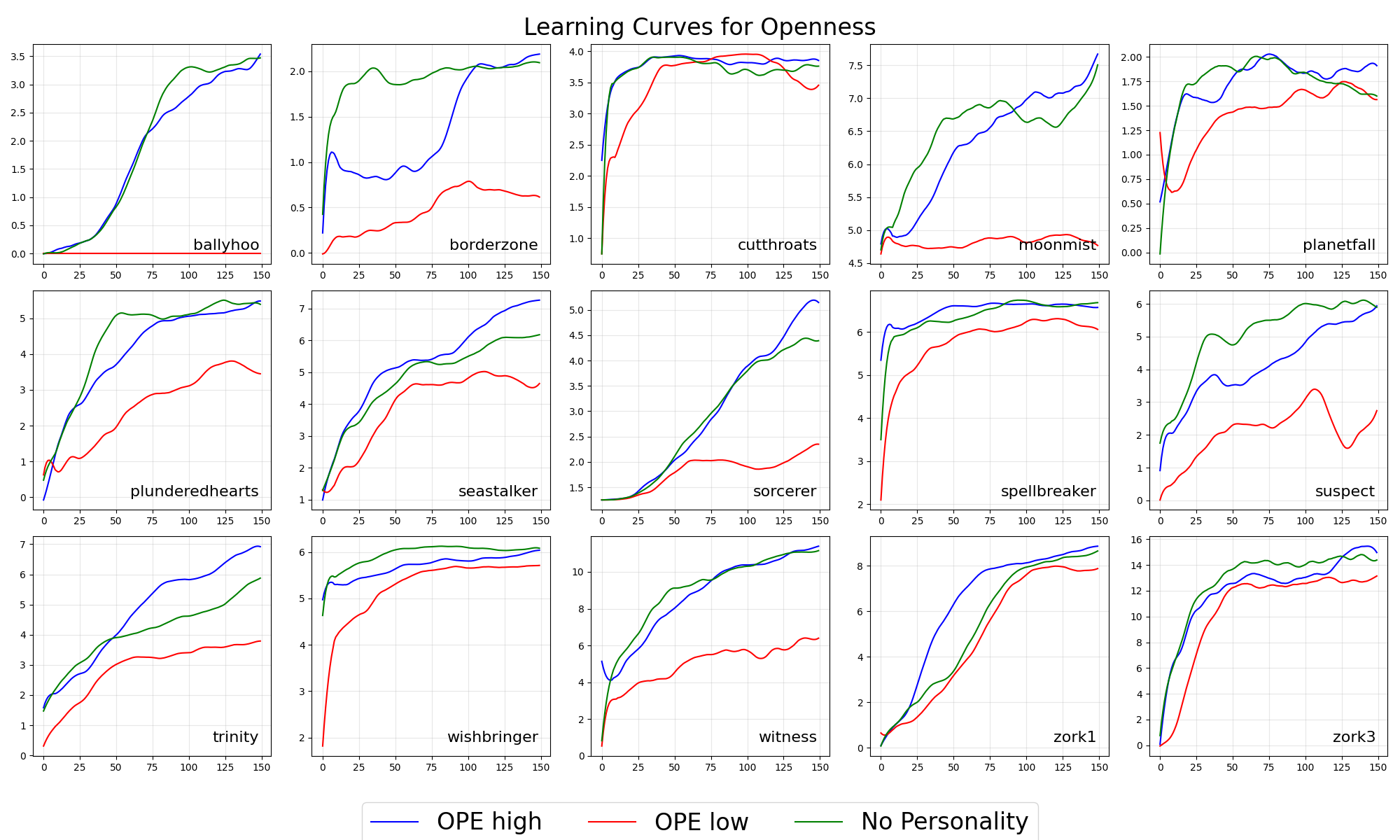}
    \caption{Learning curve comparison between \agentopeh, \agentopel\ and \agentnp\ on 15 games in Jiminy Cricket benchmark. Scores are reported at intervals of 100 training steps. Full Results are in Appendix~\ref{sec:learningfull}.}
    \label{fig:learning_curve}
    \vspace{-0.5em}
\end{figure*}

Figure~\ref{fig:learning_curve} shows the learning curves of \agentopeh, \agentnp, and \agentopel\ across 15 games to demonstrate score progression during reinforcement learning. When compared with agents of different personality types, 1) \agentopeh learns to achieve higher scores than \agentnp in numerous games, and 2) exhibits substantial gap from \agentopel's learning curve throughout the learning process. Learning curves for agents with other personality types are in Figure~\ref{fig:learning_curve_full_start} to~\ref{fig:learning_curve_full_end}.

\section{Additional Analysis}\label{sec:other-analysis}
\subsection{Impact of Psychopathy}
After analyzing and comparing the trajectories of game agents with walkthrough trajectories, we discovered that in some games, antisocial behavior is necessary to achieve higher game scores. For example, in the game ``Zork 1'', the player should kill the troll to pass through the troll room by performing actions such as ``kill troll with sword''. 

We hypothesized that among the dark triad personality types, agents with the \textbf{Psychopathy} trait would display different behavioral patterns in these situations. To test this, we filtered high-psychopathy actions from \S\ref{sec:walkthrough_analysis}, then investigated the probability of those actions performed by game agents of each personality type among the generated action candidates~\footnote{Since analyzing experimental trajectories retrospectively to analyze an agent's behavior in a specific state involves very few samples and is probabilistic, we verified action value by feeding specific states and action candidates to the policy network after RL learning was completed.}.

According to Table~\ref{tab:psychopathy_analysis}, \agentpsyh~is more likely to select actions with high psychopathy in situations that require such behavior during the walkthrough, exhibiting on average 14\% higher frequency compared to other agents. However, Table~\ref{tab:slam} shows that \agentpsyh~has difficulty exploring the game space compared to other agents (especially \agentopeh), therefore significant score improvement is not observed in Table~\ref{tab:jericho_score}. 

Also, \agentpsyh~is followed by \agentmach~in terms of frequency of actions with high psychopathy in Table~\ref{tab:psychopathy_analysis}, which we attribute to the established correlation between high psychopathy and high machiavellianism traits~\cite{mchoskey1998machiavellianism, ali2010investigating}.

\input{tables/psychopathy_analysis}

\subsection{Alignment with Human Gameplay}
\label{sec:human_gameplay}
To evaluate the alignment with human decision patterns across agents with different personality traits, we obtained human trajectories and investigated which personality type of agent shows the most similar action decision pattern to humans. For a given game state \(s_t\) and action candidates, we measured the concordance between the action chosen by the human players and the actions chosen by agents with different personality types. For comparison, we also measured concordance with a DRRN Agent without any assigned personality traits (\agentnp).

As shown in Table~\ref{tab:human_alignment_agent}, the behavior patterns of DRRN Agents without assigned personality traits (\agentnp) show high concordance with \agentopeh~and \agentagrh, while human players show high concordance with \agentpsyh~and \agentmach~compared to other personality types.

\input{tables/human_alignment_agent}

\subsection{Correlation between personality traits}
In the real world, each personality does not exist independently in isolation but rather exists in complex interrelationship with other personality types~\cite{van2010general}. To investigate patterns of personality correlations in personality types defined over text-based game actions, we examined the interrelationships between personality annotations of actions chosen by each agent. See Appendix~\ref{sec:correlation} for results and analysis.

\section{Related Work}

\subsection{Personality and LLMs}
Assessing personality in large language models has become an active research area recently ~\cite{miotto2022gpt, dorner2023personality}. Most studies use psychometric tests originally designed for humans, like the Big Five Inventory, or machine-generated tests. However, these self-assessment tests lack detailed scenarios and are sensitive to factors like phrasing ~\cite{song2023have, caron-srivastava-2023-manipulating, Huang_2023}, making them unreliable for evaluating model personality. \\
To address these challenges, researchers are exploring alternative methodologies for more accurately assessing the personality traits of language models. One promising direction involves using interactive scenarios where the language model's responses are evaluated by human judges or through automated sentiment analysis~\cite{gupta2024selfassessmenttestsunreliablemeasures, dorner2023personality, frisch2024llmagentsinteractionmeasuring, lee2024llms}. This approach aims to capture more nuanced aspects of personality that may be overlooked by standard self-assessment tests.

\subsection{Text-based game}

Research on text-based games has extensively investigated a wide range of reinforcement learning methodologies and system architectures, emphasizing the challenge of managing expansive, combinatorial action spaces shaped by natural language~\cite{he2016deep, narasimhan-etal-2015-language, xu2020deep, xu2022perceiving}. To overcome these challenges, research has been conducted on using language models to generate valid actions~\cite{he2016deep, hausknecht2020interactive, xu2021generalization, yao-etal-2020-keep}. Recently, there have been attempts to assign values related to morality and social norms in adventure games where exploration involves morally questionable actions~\cite{hendrycks2021moral, hendrycks2021what}. GALAD~\cite{ammanabrolu2022aligning} utilizes the Delphi~\cite{jiang2022machines} morality oracle and guides the agent toward creating an unharmful and successful game agent.

\section{Conclusion}
In this study, we introduced personality traits into text-based game agents and demonstrated that these traits can guide agent behavior and improve performance. Notably, agents with high openness explored more regions, engaged in effective interactions, and consequently achieved higher scores. This work highlights the potential of leveraging personality characteristics in agent design, paving the way for more nuanced and human-like AI decision-making agents.

\newpage
\section{Limitation}
\paragraph{Multifaceted Nature of Human Personality}
Human personality is inherently complex, characterized by the interplay and combination of multiple traits that collectively define an individual's behavior and responses. In this study, each agent was assigned a single, distinct personality trait based on the Big Five and Dark Triad frameworks. However, in reality, individuals exhibit a blend of various personality traits simultaneously, which interact in nuanced and context-dependent ways. Future research could integrate multiple traits to more accurately reflect the complexity of human personalities, thereby enhancing the development of more sophisticated and adaptable AI agents.

\section{Ethical Considerations}
\paragraph{Anthropomorphism}
Attributing human-like personality traits to artificial agents, as explored in this study, involves anthropomorphism—the attribution of human characteristics to non-human entities~\cite{airenti2015cognitive}. While our approach enhances agent interaction and performance in text-based games by simulating diverse personality traits, it is important to clarify that these agents do not possess consciousness, emotions, or subjective experiences.

Misinterpreting personality-driven behaviors may lead users to form unrealistic expectations or emotional attachments to agents, potentially resulting in ethical concerns. To prevent such issues, we emphasize that personality traits in our agents are functional attributes aimed at improving alignment with human users, rather than indicators of sentient beings~\cite{safdar2020ethical}.

\section*{Acknowledgments}
This work was partially funded by an unrestricted gift from Google.
This work was supported by National Research Foundation of Korea (NRF) grant funded by the Korea government (MSIT) (No. RS-2024-00354218), Institute of Information \& communications Technology Planning \& Evaluation (IITP) grant funded by the Korea government(MSIT) (No. RS-2024-00457882, AI Research Hub Project), Ministry of Culture, Sports and Tourism in 2024 (Project Name: Development of multimodal UX evaluation platform technology for XR spatial responsive content optimization, Project Number: RS-2024-00361757) and NCSOFT.

\bibliography{main}
\newpage
\label{sec:appendix}
\input{appendix}

\end{document}

%% file: tables/personality_abbrv.tex
\begin{table}[h]
\centering
\resizebox{\columnwidth}{!}{
\begin{tabular}{c|c|c|c}
\toprule
\textbf{Abbr.} & \textbf{Full Term} & \textbf{Abbr.} & \textbf{Full Term} \\
\midrule
Ope. & Openness & Neu. & Neuroticism \\
Con. & Conscientiousness & Psy. & Psychopathy \\
Ext. & Extraversion &  Nar. & Narcissism \\
Agr. & Agreeableness  & Mac. & Machiavellianism \\
\bottomrule
\end{tabular}
}
\caption{Abbreviations (Abbr.) and Full Terms for Personality Traits. Ope, Con, Ext, Agr and Neu are from Big-5, and Phy, Nar and Mac are from Dark Triad.}
\label{tab:abbreviations}
\end{table}

%% file: tables/classifier_performance.tex
\begingroup
\setlength{\tabcolsep}{0.8mm}  %
\begin{table}[htp!]

\resizebox{\linewidth}{!}{
\begin{tabular}{l|cccc|c}
\toprule
 & \textbf{BFI} & \textbf{SD-3} & \textbf{IPIP} & \textbf{Essays} & \textbf{Avg.} \\
\midrule
Flan-T5-small & 84.09 & 81.48 & 70.00 & 38.13 & 68.43 \\ 
Flan-T5-base & 95.45 & 92.59 & 85.33 & 42.68 & 79.01 \\ 
Flan-T5-large & \textbf{100} & \textbf{100} & \textbf{92.33} & 37.45 & 82.45 \\ 
Flan-T5-XL & \textbf{100} & 96.29 & 82.66 & \textbf{51.03} & \textbf{82.50} \\ 
GPT-4o-mini & 81.81 & 22.22 & 70.00 & 23.79 & 49.45 \\ 
\bottomrule
\end{tabular}
}
\caption{Classifier performance across diverse personality data~\cite{John1999TheBF, goldberg1999broad, jones2014introducing, pennebaker1999linguistic} and model size. GPT-4o-mini is zero-shot, and the other 4 are finetuned with our data. Random Chance is 33.3\%.}
\label{tab:classifier_performance}
\end{table}
\endgroup

%% file: tables/jericho_results.tex
\begingroup
\setlength{\tabcolsep}{1.0mm} %
\begin{table*}[h]
\centering
\small
\resizebox{\linewidth}{!}{
\begin{tabular}{p{0.8cm}|c|cc|cc|cc|cc|cc|cc|cc|cc}
\toprule
\multirow{2}{*}{\textbf{Game}} & \multirow{2}{*}{\agentnp}& \multicolumn{2}{c}{\textbf{Ope.}} & \multicolumn{2}{c}{\textbf{Con.}} & \multicolumn{2}{c}{\textbf{Ext.}} & \multicolumn{2}{c}{\textbf{Agr.}} & \multicolumn{2}{c}{\textbf{Neu.}} & \multicolumn{2}{c}{\textbf{Psy.}} & \multicolumn{2}{c}{\textbf{Mac.}} & \multicolumn{2}{c}{\textbf{Nar.}} \\ 
\cmidrule{3-18}
& & \agentopeh & \agentopel &
\agentconh & \agentconl &
\agentexth & \agentextl &
\agentagrh & \agentagrl &
\agentneuh & \agentneul &
\agentpsyh & \agentpsyl &
\agentmach & \agentmacl &
\agentnarh & \agentnarl\\
\midrule

BAL &3.4 &\textbf{3.5} &0.0 &2.6 &2.7 &2.9 &2.9 &3.2 &3.4 &\textbf{3.5} &2.6 &2.9 &2.5 &3.2 &3.2 &1.9 &1.5 \\

BOR &1.9 &\textbf{2.2} &0.6 &\textbf{2.0} &\textbf{1.4} &1.4 &0.6 &0.9 &0.6 &\textbf{2.0} &1.8 &1.8 &1.3 &1.1 &0.7 &1.9 &0.6 \\

CUT &3.9 &3.9 &3.4 &3.7 &3.8 &3.9 &3.8 &3.8 &3.8 &\textbf{3.9} &3.6 &3.8 &3.9 &\textbf{3.9} &\textbf{3.7} &3.7 &\textbf{3.9} \\

MOO &6.9 &\textbf{7.6} &4.8 &6.9 &6.1 &6.2 &4.9 &\textbf{7.2} &5.5 &7.0 &7.9 &\textbf{8.2} &5.8 &7.5 &7.6 &6.6 &5.4 \\

PLA &1.8 &\textbf{1.9} &\textbf{1.6} &1.7 &1.8 &1.7 &1.7 &\textbf{1.8} &1.8 &\textbf{1.7} &1.7 &\textbf{1.9} &1.7 &1.8 &1.7 &1.7 &1.7 \\

PLU &5.3 &\textbf{5.5} &3.5 &5.0 &3.6 &4.4 &4.8 &5.3 &4.3 &\textbf{5.4} &4.9 &5.2 &4.2 &5.1 &3.9 &5.3 &3.8 \\

SEA &5.1 &\textbf{7.3} &4.6 &5.8 &5.7 &6.3 &5.9 &6.6 &6.0 &6.0 &5.7 &\textbf{6.1} &5.0 &6.2 &6.1 &6.4 &6.9 \\

SOR &3.8 &\textbf{5.2} &2.4 &\textbf{4.5} &3.0 &\textbf{4.4} &3.0 &4.5 &4.1 &3.1 &3.4 &\textbf{4.4} &3.0 &4.5 &4.4 &\textbf{4.2} &2.2 \\

SPE &6.6 &6.6 &6.1 &6.4 &5.0 &\textbf{6.8} &6.3 &6.6 &6.5 &5.6 &6.5 &6.5 &6.5 &6.6 &6.2 &6.4 &5.1 \\

SUS &4.6 &\textbf{5.9} &2.7 &4.5 &3.9 &4.4 &4.1 &4.1 &3.0 &5.2 &5.1 &\textbf{5.2} &3.0 &3.3 &4.5 &\textbf{5.2} &2.8 \\

TRI &4.0 &\textbf{6.9} &3.8 &6.3 &5.4 &5.6 &6.6 &5.6 &6.6 &5.0 &5.9 &6.0 &4.7 &6.2 &5.7 &5.2 &6.1 \\

WIS &6.1 &6.0 &5.7 &5.8 &5.8 &5.9 &5.8 &5.8 &5.8 &\textbf{6.2} &6.0 &6.1 &6.1 &5.9 &5.8 &6.0 &5.9 \\

WIT &10.9 &\textbf{11.4} &6.4 &10.6 &6.5 &8.3 &9.7 &9.2 &9.1 &\textbf{11.1} &10.6 &\textbf{11.3} &7.1 &10.2 &8.8 &\textbf{11.1} &8.5 \\

Z1 &6.8 &8.9 &7.9 &8.8 &8.5 &8.3 &9.0 &8.7 &8.8 &7.8 &9.0 &8.3 &8.7 &8.4 &7.1 &8.8 &8.6 \\

Z3 &13.3 &\textbf{15.0} &13.1 &13.0 &13.2 &13.0 &\textbf{14.6} &\textbf{14.3} &12.6 &13.1 &\textbf{14.7} &13.8 &14.1 &14.9 &13.9 &13.8 &14.1 \\

\midrule
\textbf{Avg.} &5.6 &\textbf{6.5} &4.4 &5.8 &5.1 &5.6 &5.7 &5.8 &5.4 &5.8 &6.0 &\underline{6.0} &5.3 &5.9 &5.5 &5.7 &5.3 \\
\midrule
\textbf{Cnt.} & - & \textbf{11} & 0 & 2 & 1 & 2 & 1 & 4 & 0 & \underline{6} & 2 & 5 & 2 & 1 & 0 & 2 & 2 \\

\midrule
\textbf{Diff.} & - & \multicolumn{2}{c|}{\textbf{+2.1}} & \multicolumn{2}{c|}{+0.7} & \multicolumn{2}{c|}{-0.1} & \multicolumn{2}{c|}{+0.4} & \multicolumn{2}{c|}{-0.2} & \multicolumn{2}{c|}{\underline{+0.7}} & \multicolumn{2}{c|}{+0.4} & \multicolumn{2}{c}{+0.4} \\
\bottomrule

\end{tabular}
}
\caption{Game Scores on games of Jiminy Cricket. \agentnp (`No Personality') means no guidance with personality classifier, and the symbols (\textuparrow) and (\textdownarrow) indicate high and low levels of each personality trait, respectively. \textbf{Avg.}, \textbf{Cnt.} and \textbf{Diff.} are three criteria defined in \S\ref{sec:jmn-result}. We only report 15 games here because in the remaining 10 games, agents of any personality type failed to score points in over 90 percent of attempts. Results for all games can be found in Table~\ref{tab:unnamed_table_1} and~\ref{tab:unnamed_table_2}.
For scores, \textbf{bold} indicates games satisfying the threshold condition for \textbf{Cnt.} The best scores are \textbf{bolded} and the second-best ones are \underline{underlined} on metrics.}

\label{tab:jericho_score}
\end{table*}
\endgroup

%% file: tables/jericho_stat_2.tex
\begin{table*}[!htp]\centering
\resizebox{\linewidth}{!}{
\begin{tabular}{c|c|c|cccccccc}\toprule
\textbf{Test Type}& \textbf{Comparison Pair}& \textbf{Stat.} &\textbf{Ope.} &\textbf{Con.} &\textbf{Ext.} &\textbf{Agr.} &\textbf{Neu.} &\textbf{Psy.} &\textbf{Mac.} &\textbf{Nar.} \\

\midrule
\midrule

\multirow{6}{*}{\parbox{2cm}{\centering \textit{Wilcoxon}  \\ \textit{Signed Rank  (\textdownarrow)}}}
&\multirow{2}{*}{(\agenthigh, \agentnp)} &\textit{T} &\textbf{1.0} &50.5 &48.0 &29.5 &34.0 &19.0 &28.0 &29.5 \\
\cline{3-3}
& &\textit{p}-value &\textbf{0.002} &0.900 &0.777 &0.456 &0.244 &0.035 &0.388 &0.263 \\
\cline{2-11}
&\multirow{2}{*}{(\agentnp, \agentlow)} &\textit{T} &\textbf{8.0} &28.0 &49.0 &34.5 &47.0 &23.0 &57.5 &36.5 \\
\cline{3-3}
& &\textit{p}-value &\textbf{0.002} &0.124 &0.561 &0.442 &0.489 &0.116 &0.934 &0.315 \\
\cline{2-11}
&\multirow{2}{*}{(\agenthigh, \agentlow)} &\textit{T} &\textbf{0.0} &13.5 &44.5 &15.5 &40.0 &8.0 &13.5 &19.0 \\
\cline{3-3}
& &\textit{p}-value &\textbf{0.000} &0.014 &0.944 &0.065 &0.432 &0.009 &0.014 &0.035 \\

\cline{1-11}

\multirow{2}{*}{\parbox{2cm}{\centering \textit{Friedman  (\textuparrow)}}}
&\multirow{2}{*}{(\agenthigh, \agentnp, \agentlow)} &\textit{Fr} &\textbf{25.2} & 6.0 & 3.9 & 3.6 & 1.7 & 6.0 & 5.0 & 3.9 \\
\cline{3-3}
& &\textit{p}-value &\textbf{0.000} & 0.049 & 0.143 & 0.168 & 0.430 & 0.049 & 0.084 & 0.143 \\

\bottomrule
\end{tabular}
}
\caption{Statistical analysis of all scores shown in Table~\ref{tab:jericho_score}. \agenthigh, \agentnp, \agentlow denotes each groups consisting of scores from 15 games (n=15). \textit{T} and \textit{Fr} denotes the test statistic, and the \textit{p}-value denotes the significance probability of each test. The best scores are \textbf{bolded} and the second-best ones are \underline{underlined}.}\label{tab:jericho_stat}
\end{table*}

%% file: tables/slam.tex
\begingroup
\setlength{\tabcolsep}{1.0mm} %
\begin{table*}[h]
\centering
\small
\resizebox{\linewidth}{!}{
\begin{tabular}{c|c|ccccccccccc}
\toprule
\textbf{Metric}& &\agentnp &\agentopeh &\agentopel &\agentconh &\agentexth &\agentagrh &\agentneuh &\agentpsyh &\agentnarh &\agentmach \\
\midrule
\textit{Trajectory Length (\textuparrow)} & - & 45.85&\underline{57.04} &39.86 &50.05 &\textbf{60.91} &49.17 &48.85 &50.38 &48.71 &46.07 \\
\cline{1-12}
\multirow{2}{*}{\rule[-0.1ex]{0pt}{5.1ex} \textit{Visit Count (\textuparrow)}} & \rule[-1ex]{0pt}{3.7ex} \textit{Com.} &8.66 & \textbf{8.96} & 8.02 & \underline{8.88} & 7.83 & 8.55 & 8.88 & 8.29 & 8.65 & 8.07 \\
\rule[-1ex]{0pt}{3.5ex} &\textit{Unc.} &0.83 &\underline{1.20} &0.30 &0.89 &0.88 &0.67 &\textbf{1.21} &0.82 &1.01 &0.64 \\
\cline{2-12}
\rule[-1ex]{0pt}{3.5ex} &\textit{Total.} & 9.49 & \textbf{10.16} & 8.32 & 9.77 & 8.71 & 9.22 & \underline{10.09} & 9.11 & 9.66 & 8.71 \\

\cline{1-12}
\multirow{2}{*}{\rule[-0.1ex]{0pt}{4.9ex} \textit{Avg. Step (\textdownarrow)}} & \rule[-1ex]{0pt}{3.7ex} \textit{Com.} &12.64 &\underline{11.93} &11.60 &14.45 &\textbf{10.34} &12.3 &13.84 &13.35 &12.03 &12.37 \\
\rule[-0.6ex]{0pt}{3.1ex} &\textit{Unc.} & 8.62 & 6.39 & 12.01 & 17.54 & 6.15 & 9.36 & 16.90 & 8.52 & 9.81 & 8.87\\
\cline{2-12}
\rule[-1ex]{0pt}{3.5ex} &\textit{Total.} & 21.26 & \underline{18.32} & 23.61 & 31.99 & \textbf{16.49} & 21.66 & 30.74 & 21.87 & 21.84 & 21.24 \\

\bottomrule
\end{tabular}
}
\caption{Analysis in last 50 Episodes based on each game agent's movement trajectory in Zork1. Standard deviations and scores of omitted agents are provided in Table~\ref{tab:slam_detail1} and ~\ref{tab:slam_detail2}. The best scores are \textbf{bolded} and the second-best ones are \underline{underlined}.}

\label{tab:slam}
\end{table*}
\endgroup

%% file: tables/qvalue.tex
\begingroup
\setlength{\tabcolsep}{2mm}

\begin{table}[htp!]
\centering
\resizebox{\columnwidth}{!}{

\begin{tabular}{c|cc|c|cc}
\toprule
\renewcommand{\arraystretch}{10}

\textbf{Agent} & \( Q(s_t, a_t) \) & \textit{Cnt.} & \textbf{Agent} & \( Q(s_t, a_t) \) & \textit{Cnt.}  \\ 
 
\midrule
\agentopeh & \textbf{18.3} & \textbf{6.2} & \agentopel & 12.7 & 3.1 \\
\agentconh & 14.7 & 3.7 & \agentconl & 13.4 & 3.2 \\
\agentexth & 15.2 & 5.4 & \agentextl & 13.9 & 4.8 \\
\agentagrh & 14.8 & 4.8 & \agentagrl & 14.2 & 3.9 \\
\agentneuh & 16.4 & 3.1 & \agentneul & 13.1 & 3.4 \\
\agentpsyh & 15.9 & 4.9 & \agentpsyl & 12.8 & 3.1 \\
\agentnarh & 15.1 & 3.4 & \agentnarl & 14.6 & 4.3 \\
\agentmach & 13.5 & 3.6 & \agentmacl & 14.3 & 3.6 \\
\bottomrule
\end{tabular}
}

\caption{Analysis of reward-earning actions performed by each agent.
\(Q\) denotes the action value from each agent's policy network for \( (s_t, a_t) \)  where reward was given, and \textit{Cnt.} denotes the number of reward-earning actions performed by each agent within a single episode. Each score is the average over the last 50 episodes.}

\label{tab:qvalue}
\end{table}
\endgroup

%% file: tables/action_ratio.tex
\begingroup
\setlength{\tabcolsep}{2mm}

\begin{table}[htp!]
\centering
\resizebox{\columnwidth}{!}{

\begin{tabular}{c|cc|c|cc}
\toprule
\renewcommand{\arraystretch}{10}
\multirow{2}{*}{\rule[-1ex]{0pt}{3.7ex}\textbf{Agent}} & \multicolumn{2}{c|}{\rule[-1.3ex]{0pt}{3.5ex}$r(a_{p\uparrow}) - r(a_{p\downarrow})$} & \multirow{2}{*}{\textbf{Agent}} & \multicolumn{2}{c}{$r(a_{p\uparrow}) - r(a_{p\downarrow})$} \\
\cline{2-3} \cline{5-6}
 & \rule[-1ex]{0pt}{3.7ex}\textit{Init50} & \textit{Fin50} & & \textit{Init50} & \textit{Fin50} \\ 
 
\midrule
\agentopeh & 0.45 & \textbf{0.58} & \agentopel & -1.00 & -0.17 \\
\agentconh & 0.40 & 0.22 & \agentconl & -0.83 & -0.66 \\
\agentexth & 0.38 & 0.25 & \agentextl & -0.52 & -0.33 \\
\agentagrh & 0.48 & 0.28 & \agentagrl & -0.66 & -0.57 \\
\agentneuh & 0.65 & 0.53 & \agentneul & -0.26 & \textbf{-0.33} \\
\agentpsyh & -0.18 & -0.31 & \agentpsyl & -0.81 & -0.58 \\
\agentnarh & 0.17 & \textbf{0.31} & \agentnarl & -0.65 & \textbf{-0.66} \\
\agentmach & 0.44 & 0.02 & \agentmacl & -0.88 & -0.49 \\
\bottomrule
\end{tabular}
}

\caption{
The difference between the normalized ratios of \actionhigh\ and \actionlow\ performed by each agent with different personalities during training. \textit{Init50} and \textit{Fin50} denote the first and last 50 episodes of each training process, respectively. \textbf{Bold} indicates agents with increased ratios for \agenthigh\ and decreased for \agentlow.}

\label{tab:action_ratio}
\end{table}
\endgroup

%% file: tables/personality_walkthrough.tex
\begingroup
\setlength{\tabcolsep}{1.1mm}
\begin{table}[htp!]
\centering
\resizebox{\columnwidth}{!}{
\begin{tabular}{l|cccccccc}
\toprule
 & Ope. & Con. & Ext. & Agr. & Neu. & Psy. & Mac. & Nar. \\
\midrule
\actionhigh & \textbf{18.6} & 12.6 & 3.8 & 10.6 & 2.7 & 6.6 & 4.2 & 6.9 \\
\actionlow & 2.1 &15.1 & 1.8 & 12.6 & 0.0 & 0.7 & 1.2 & 0.4 \\
\bottomrule
\end{tabular}
}
\caption{Analysis of personality traits in walkthrough actions for 25 games. Numbers~(\%) represent the ratio of each actions among all types of personality.}
\label{tab:walkthrough}
\end{table}
\endgroup

%% file: tables/psychopathy_analysis.tex
\begingroup
\setlength{\tabcolsep}{2mm}

\begin{table}[htp!]
\centering
\resizebox{\columnwidth}{!}{
\begin{tabular}{c|c|c|c}
\toprule
\renewcommand{\arraystretch}{10}

\textbf{Agent} & \( P(a_c = a_{\text{Psy},\uparrow}) \) & 
\textbf{Agent} & \( P(a_c = a_{\text{Psy},\uparrow}) \)  \\ 
 
\midrule
\agentopeh & 12.1 & \agentopel & 14.4  \\
\agentconh & 12.9 & \agentconl & 12.9  \\
\agentexth & 18.1 & \agentextl & 1.1  \\
\agentagrh & 9.8 & \agentagrl & 18.4  \\
\agentneuh & 10.8 & \agentneul & 9.2  \\
\agentpsyh & \textbf{25.5} & \agentpsyl & 1.4  \\
\agentmach & \underline{20.7} & \agentmacl & 13.0 \\
\agentnarh & 13.6 & \agentnarl & 2.3  \\
\bottomrule
\end{tabular}
}

\caption{Analysis for actions demonstrating high psychopathy tendencies. With 10 action candidates generated by the action generator, \( P(a_c = a_{\text{Psy},\uparrow}) \) denotes the probability that \( a_{\text{Psy},\uparrow}  \) to be selected based on the softmax distribution of action values assigned by each agent's trained policy network.}
\label{tab:psychopathy_analysis}
\vspace{-5mm}
\end{table}
\endgroup

%% file: tables/human_alignment_agent.tex
\begingroup
\setlength{\tabcolsep}{1.1mm}
\begin{table}[htp!]
\centering
\resizebox{\columnwidth}{!}{
\begin{tabular}{l|ccccccccc}
\toprule
Agent &  & Ope. & Con. & Ext. & Agr. & Neu. & Psy. & Mac. & Nar. \\
\midrule
\multirow{2}{*}{Human} 
& \agenthigh & 11.9 & 11.8 & 18.5 & 7.1 & 11.2 & \textbf{28.8} & \textbf{19.9} & 11.6 \\
& \agentlow & 9.7 & 10.8 & 1.4 & 11.4 & 6.1 & 1.4 & 11.0 & 1.6 \\

\midrule[0.75pt]
\addlinespace[5mm] %
\midrule
Agent &  & Ope. & Con. & Ext. & Agr. & Neu. & Psy. & Mac. & Nar. \\
\midrule

\multirow{2}{*}{\agentnp} 
& \agenthigh & \textbf{18.8} & 10.4 & 9.2 & \textbf{14.8} & 11.5 & 9.6 & 12.0 & 12.1\\ 
& \agentlow &  11.6 & 9.7 & 14.6 & 10.4 & 9.7 & 11.6 & 10.3 & 8.1\\

\bottomrule
\end{tabular}
}
\caption{
Concordance rates between 16 distinct personality type agents and both human players and baseline DRRN agents, demonstrating differential behavior patterns across personality types.}
\label{tab:human_alignment_agent}
\vspace{-5mm}
\end{table}
\endgroup

%% file: appendix.tex
\clearpage

\appendix

\section{Overview}
We provide the following details in this appendix:
\begin{itemize}
    \item In Appendix B, we provide the detailed hyperparameters applied to train the DRRN agent and personality classifier used in our framework .
    \item In Appendix C, we provide information and various examples of games of Jiminy Cricket benchmark.
    \item In Appendix D, we present the full results of results in main page, along with additional materials and analyses that were not covered on the main page.
    \item In Appendix E, We introduce the personality framework and questionnaires used in data generation.
    \item In Appendix F, we provide a detailed analysis of the creation process, as well as composition of the dataset used to train the personality classifier.

\end{itemize}

\section{Training Details}
\subsection{Details for training DRRN Agent}
Table~\ref{hyperparameters_drrn} provides the specific hyperparameters utilized for training the policy network employed in DRRN. It took up to 12 hours to complete learning for running a single game once using an NVIDIA A6000.

\begingroup
\setlength{\tabcolsep}{1mm}
\begin{table}[ht]
\centering
\begin{tabular}{ll}
\toprule
\textbf{Hyperparameter type} & \textbf{Value} \\
\midrule
\multicolumn{2}{l}{\textbf{RL Training}} \\
\midrule
Discount $\gamma$ & 0.9 \\
Replay priority & 0.5 \\
Replay buffer size & 10000 \\
Policy shaping condition weight & 2 \\
Batch size & 64 \\
Gradient clip & 5.0 \\
Steps per episode & 100 \\
Max. steps per start & 15000 \\
early stopping steps & 5000 \\
Parallel Environments & 8 \\
\midrule
\multicolumn{2}{l}{\textbf{Policy network}} \\
\midrule
Q-network feedforward size & 128 \\
GRU hidden size & 128 \\
\midrule
\end{tabular}
\caption{Hyperparameter values for RL training and policy network.}
\label{hyperparameters_drrn}
\end{table}
\endgroup

\subsection{Details for training Personality classifier}
Finetuning models for personality classification (Flan-T5-small, Flan-T5-small, Flan-T5-large, and Flan-T5-XL) took up to 24 hours, when using four NVIDIA RTX-3090s. In Table~\ref{hyperparameters_classifier}, we detail the key parameters during training.

\begingroup
\setlength{\tabcolsep}{1mm}
\begin{table}[ht]
\centering
\begin{tabular}{ll}
\toprule
\textbf{Hyperparameter type} & \textbf{Value} \\
\midrule
Learning Rate & $3e-4$ \\
Weight Decay & 0.1 \\
Adam $\beta1$ & 0.9 \\
Adam $\beta2$ & 0.95 \\
Adam $\epsilon$ & $1e-5$ \\
Training Epochs & 3 \\
Split & 0.9 \\
Split Seed & 42 \\
Early Stopping Patients & 10 \\
\toprule
\end{tabular}
\caption{Hyperparameter values for training personality classifier.}
\label{hyperparameters_classifier}

\end{table}
\endgroup

\section{Games in Jiminy Cricket environment}
\subsection{Abbreviations}
In Table~\ref{tab:abbreviation_jericho}, we denote the abbreviation of subgames in game environment.

\begin{table}[htp!]
\centering

\resizebox{\columnwidth}{!}{
\begin{tabular}{rcrc}
\toprule
\textbf{Abbr.} & \textbf{Full Term} & \textbf{Abbr.} & \textbf{Full Term} \\
\midrule
BAL & Ballyhoo & MOO & Moonmist \\
BOR & Borderzone & PLA & Planetfall \\
CUT & Cutthroats & PLU & Plunderedhearts \\
DEA & Deadline & SEA & Seastalker \\
ENC & Enchanter & SOR & Sorcerer \\
HIT & Hitchhiker & SPE & Spellbreaker \\
HOL & Hollywoodhijinx & STA & Starcross \\
INF & Infidel & STF & Stationfall \\
LUR & Lurkinghorror & SUS & Suspect \\
SUSP & Suspended & TRI & Trinity \\
WIS & Wishbringer & Z1 & Zork1 \\
WIT & Witness & Z2 & Zork2 \\
& & Z3 & Zork3 \\
\bottomrule
\end{tabular}}
\caption{Abbreviations for games in Jiminy Cricket Benchmark.}
\label{tab:abbreviation_jericho}
\end{table}

\subsection{Example of Jiminy Cricket Benchmark.}
We provide a portion of the transcript obtained from playing one of the games in the Jiminy Cricket benchmark. All of the agents playing the game are granted high openness.

\begin{tcolorbox}[colframe=black,colback=white,boxrule=0.1mm,arc=3mm,left=2mm,right=2mm,top=2mm,bottom=2mm]
    \small
    
    \textbf{Actions Candidates:} \\
    take all, take egg, take feather, throw egg at nest, get it, take nest, examine clasp, put egg in nest, get feather, climb nest, get clasp, north, inventory, examine hatch, examine gold, open hatch, down, "examine birds nest", take clasp, examine nest, take all from nest, up, examine necklace, take bird, eat nest, examine feather, examine bracelet, south, take jewel, take gold, take necklace, examine egg, take hatch, break egg, examine jewel \\
    \textbf{Observation:} \\
    Up a Tree You are about 10 feet above the ground nestled among some large branches. The nearest branch above you is above your reach. Beside you on the branch is a small bird's nest. In the bird's nest is a large egg encrusted with precious jewels, apparently scavenged by a childless songbird. The egg is covered with fine gold inlay, and ornamented in lapis lazuli and mother-of-pearl. Unlike most eggs, this one is hinged and closed with a delicate looking clasp. Egg appears extremely fragile. \\
    \textbf{Action:} get it \\
    \textbf{Score:} 4
\end{tcolorbox}

\begin{tcolorbox}[colframe=black,colback=white,boxrule=0.1mm,arc=3mm,left=2mm,right=2mm,top=2mm,bottom=2mm]
    \small
    
    \textbf{Actions Candidates:} \\
    east, hit painting, open painting, examine art, enter painting, open door, examine mural, turn diamond, get painting, climb painting, take painting, southwest, break window, close gallery, south, move painting, northeast, unlock door with key, read painting, push painting, examine paintings, west, southeast, break painting, exits, turn ring, north, examine painting, up, inventory, listen, search painting, examine portrait, northwest, out, down, look under painting, look behind painting, cut painting \\
    \textbf{Observation:} \\
    Gallery This is an art gallery. Most of the paintings have been stolen by vandals with exceptional taste. The vandals left through either the north or west exits. Fortunately, there is still one chance for you to be a vandal, for on the far wall is a painting of unparalleled beauty. \\
    \textbf{Action:} read painting \\
    \textbf{Score:} 4
\end{tcolorbox}

\begin{tcolorbox}[colframe=black,colback=white,boxrule=0.1mm,arc=3mm,left=2mm,right=2mm,top=2mm,bottom=2mm]
    \small
    
    \textbf{Actions Candidates:} \\
    look through window, southwest, out, climb tree, break window, down, examine me, look in it, north, close window, west, look in window, southeast, look, enter window, search house, knock on window, look out window, northeast, examine window, examine door, examine tree, wait, enter house, in, south, go through window, exit, examine house, knock on door, inventory, examine bed, east, enter cabin, up, look through it, again, northwest \\
    \textbf{Observation:} \\
    With great effort, you open the window far enough to allow entry. \\
    \textbf{Action:} enter window \\
    \textbf{Score:} 10
\end{tcolorbox}

\subsection{World Visualization and Trajectory of Agent}
\begin{figure*}
    \centering
    \includegraphics[width=0.8\textwidth, height=\textheight, keepaspectratio]{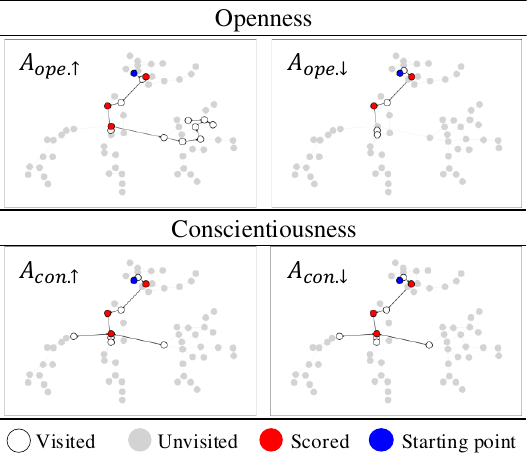}
    \caption{Trajectory of \agentopeh, \agentopel, \agentconh\ and \agentconl in Zork1.}
    \label{fig:traj_detail_1}
\end{figure*}

\begin{figure*}
    \centering
    \includegraphics[width=0.8\textwidth, height=\textheight, keepaspectratio]{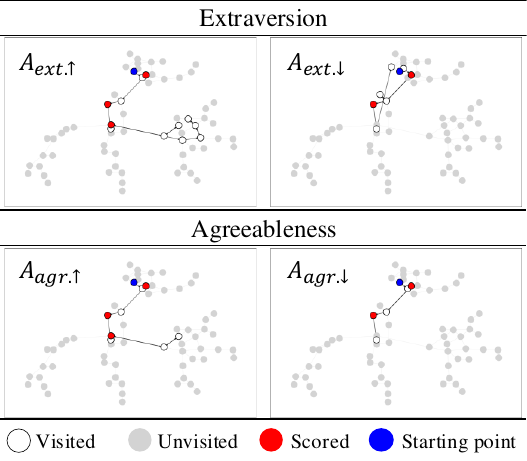}
    \caption{Trajectory of \agentexth, \agentextl, \agentagrh\ and \agentagrl in Zork1.}
    \label{fig:traj_detail_2}
\end{figure*}

\begin{figure*}
    \centering
    \includegraphics[width=0.8\textwidth, height=\textheight, keepaspectratio]{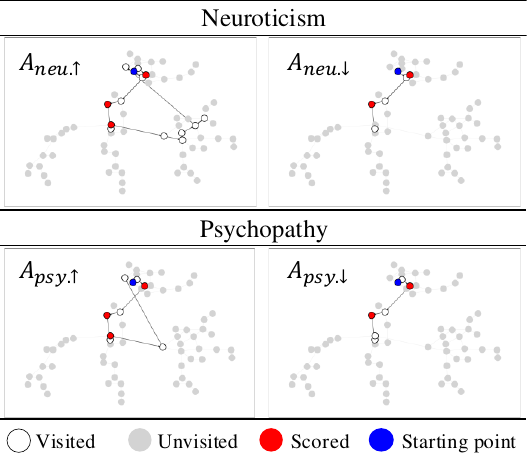}
    \caption{Trajectory of \agentneuh, \agentneul, \agentpsyh\ and \agentpsyl in Zork1.}
    \label{fig:traj_detail_3}
\end{figure*}

\begin{figure*}
    \centering
    \includegraphics[width=0.8\textwidth, height=\textheight, keepaspectratio]{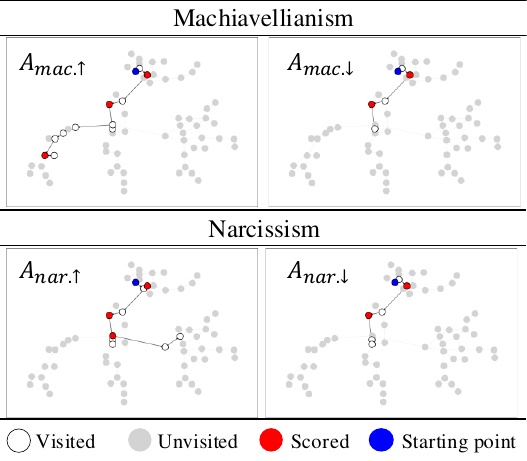}
    \caption{Trajectory of \agentmach, \agentmacl, \agentnarh\ and \agentnarl in Zork1.}
    \label{fig:traj_detail_4}
\end{figure*}

\begin{figure*}[p!]
    \centering
    \includegraphics[width=\linewidth]{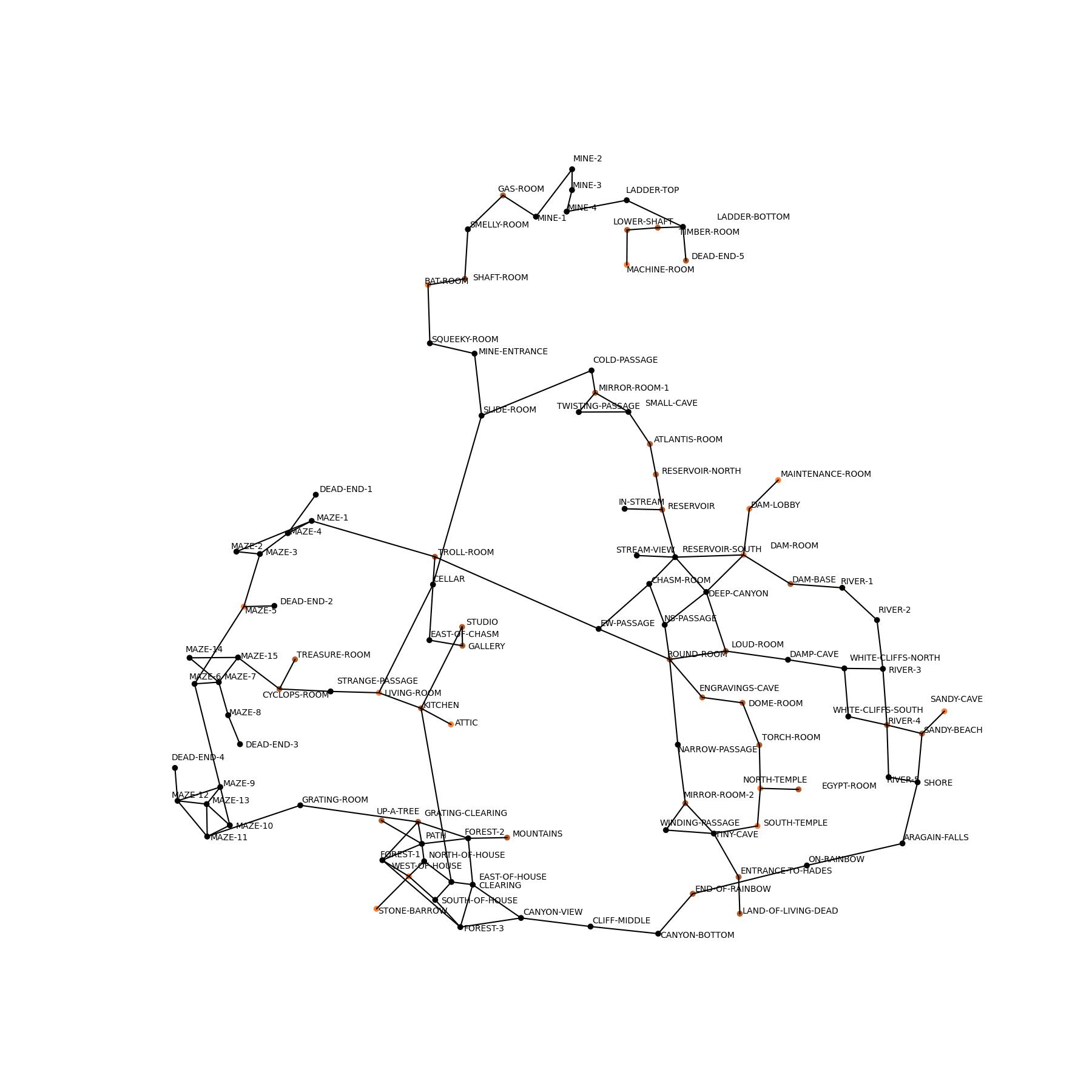}
    \caption{Visualization the entire game world - Zork1}
    \label{fig:visualizing_zork1}
\end{figure*}

\begin{figure*}[p!]
    \centering
    \includegraphics[width=\linewidth]{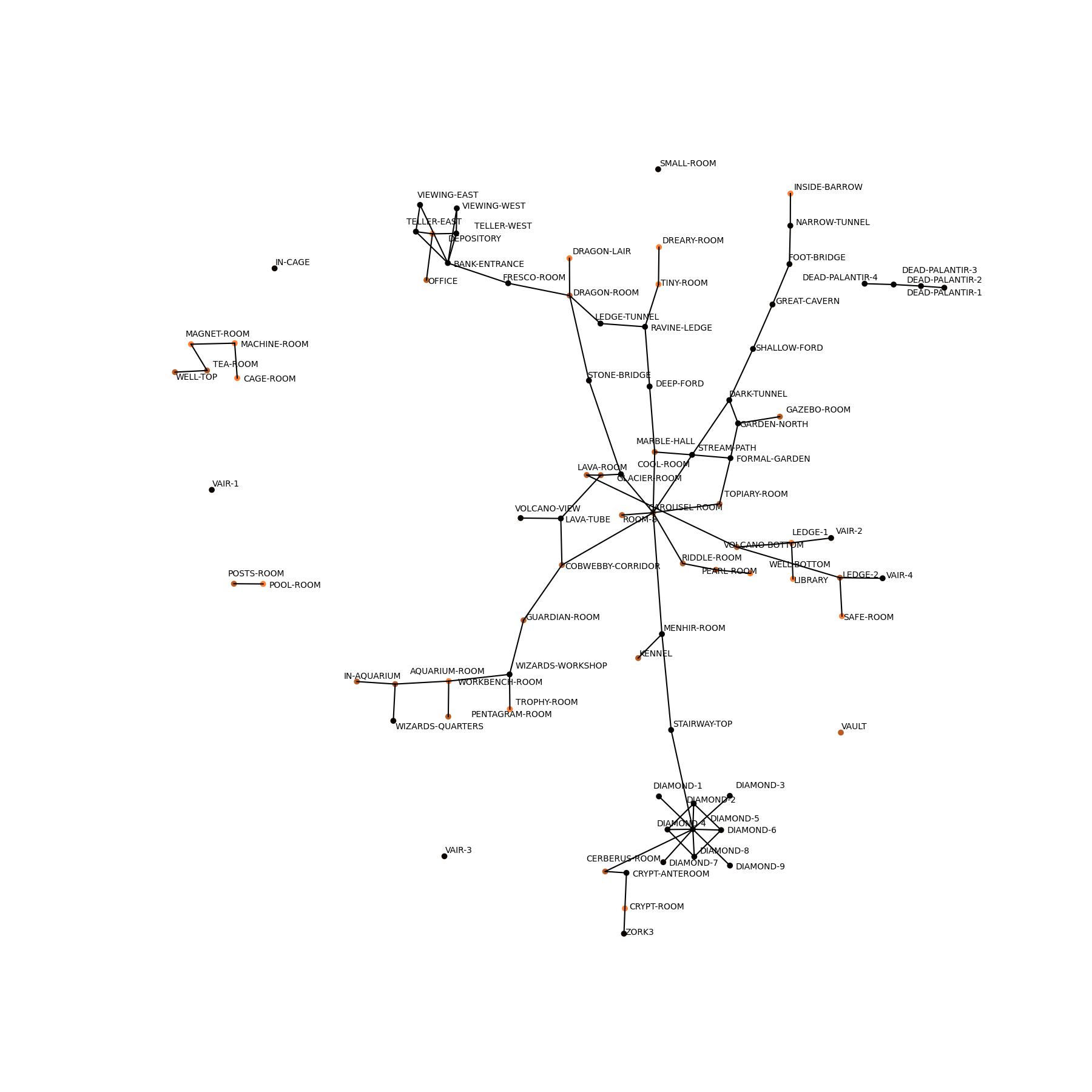}
    \caption{Visualization the entire game world - Zork2}
    \label{fig:visualizing_zork2}
\end{figure*}

\begin{figure*}[htp!]
    \centering
    \includegraphics[width=\linewidth]{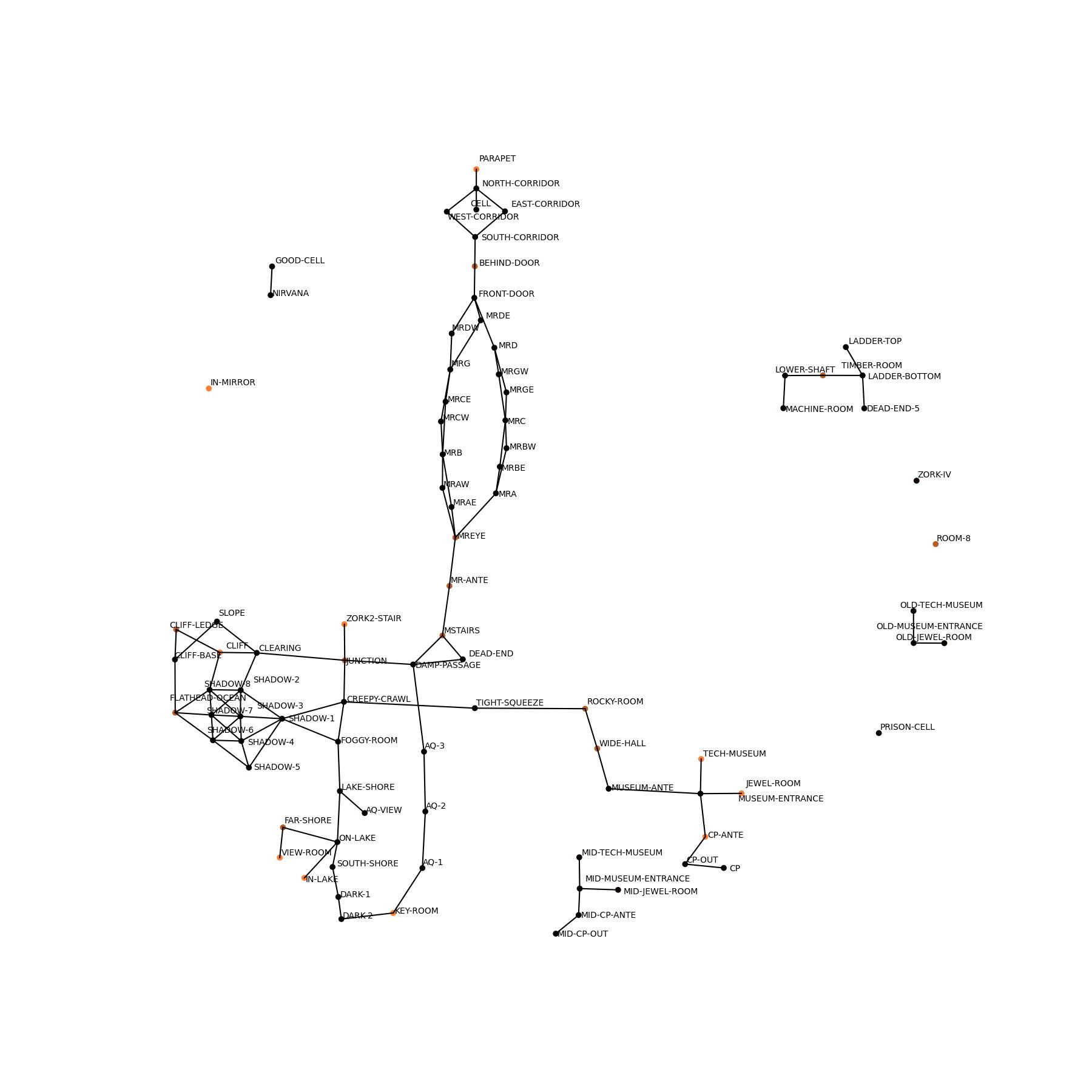}
    \caption{Visualization the entire game world - Zork3}
    \label{fig:visualizing_zork3}
\end{figure*}

We applied the visualization method presented in \S\ref{sec:traj-analysis} to all agents, with results shown in Figure \ref{fig:traj_detail_1} to \ref{fig:traj_detail_4}. Additionally, we visualize the map of the game \textit{Zork-1}, \textit{Zork-2}, and \textit{Zork-3} from Jiminy Cricket benchmark in in Figure~\ref{fig:visualizing_zork1} to~\ref{fig:visualizing_zork3}.

\section{Full Results \& Additional Analysis}

\subsection{Full Results on Table 3}
Game scores and standard deviations across three different runs for all 25 games in the Jiminy-Cricket benchmark are presented in Table~\ref{tab:unnamed_table_1} and ~\ref{tab:unnamed_table_2}.

\input{tables/unnamed_in_appendix}

\input{tables/unnamed2_in_appendix}

\subsection{Learning Curve}\label{sec:learningfull}
\begin{figure*}[htbp!]
    \centering
    \includegraphics[width=\textwidth, height=\textheight, keepaspectratio]{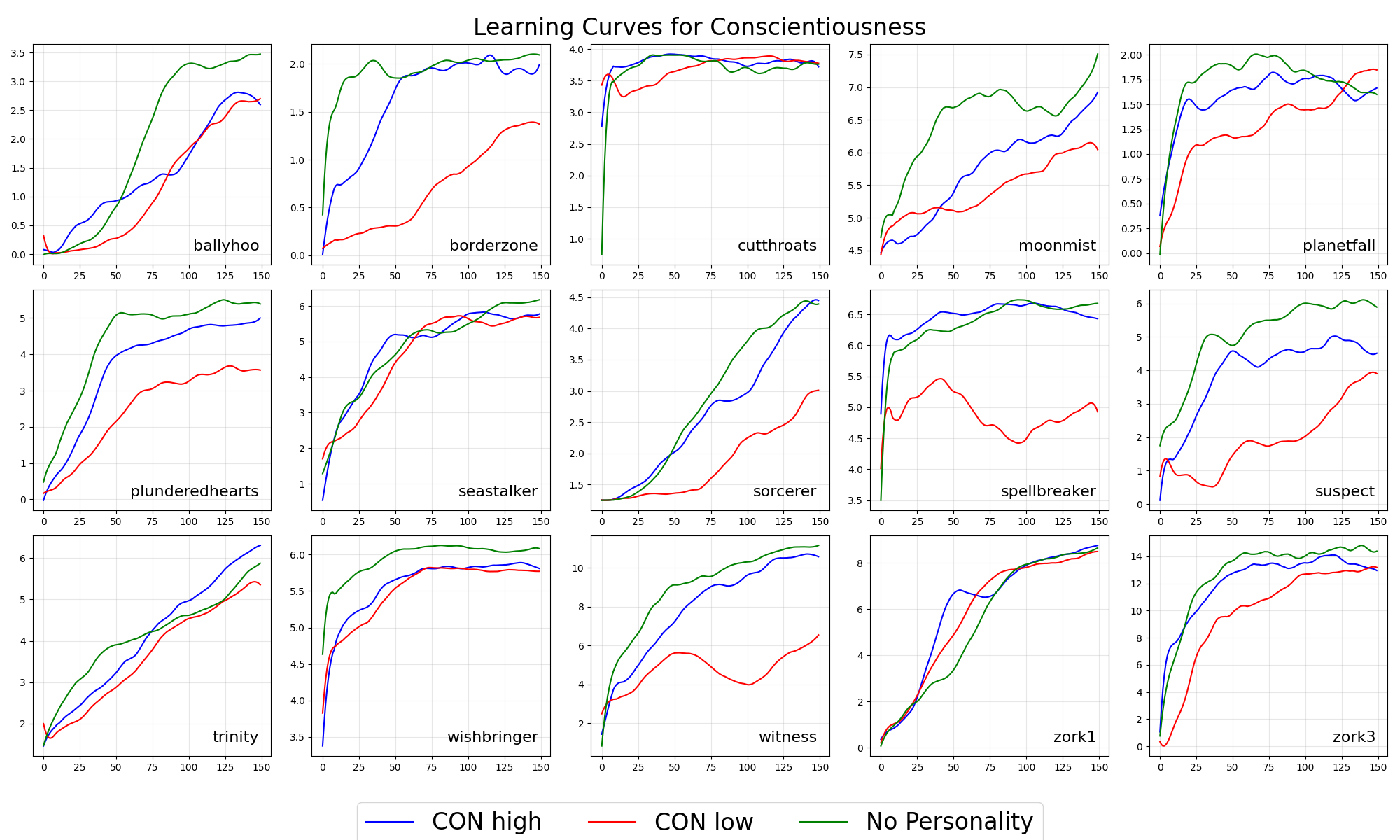}
    \caption{Learning curve each 15 games in Jiminy Cricket benchmark.}
    \label{fig:learning_curve_full_start}
\end{figure*}

\begin{figure*}[htp!]
    \centering
    \includegraphics[width=\textwidth, height=\textheight, keepaspectratio]{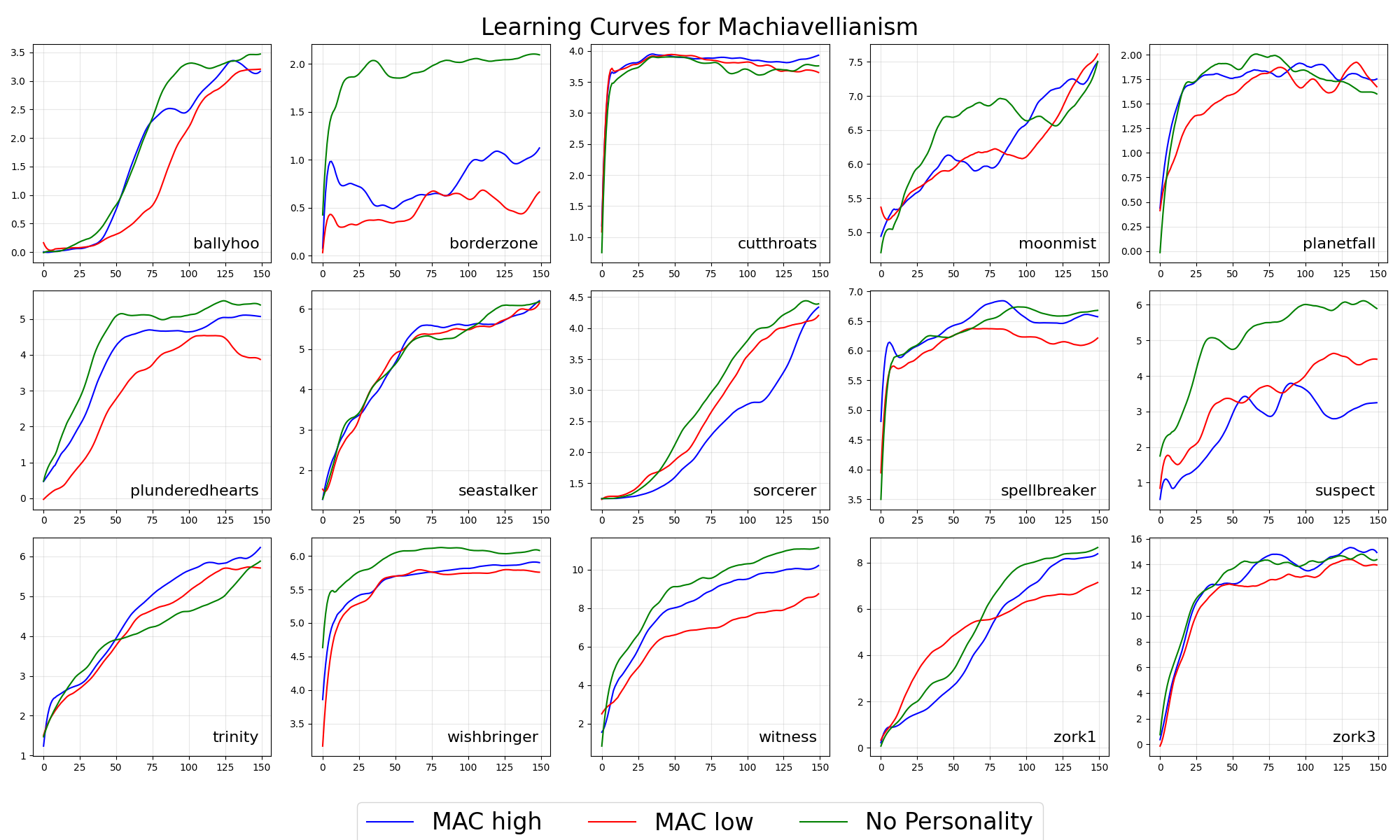}
    \caption{Learning curve each 15 games in Jiminy Cricket benchmark.}
    \label{fig:learning_curve_full_end}
\end{figure*}

We report the training progression through learning curves for all eight personality types, measured across a test suite of 15 games in Figure~\ref{fig:learning_curve_full_start} to~\ref{fig:learning_curve_full_end}.

\input{tables/room_split}

\subsection{Correlation between Personality Type}\label{sec:correlation}
\begin{figure}[t]
    \centering
    \includegraphics[width=\columnwidth]{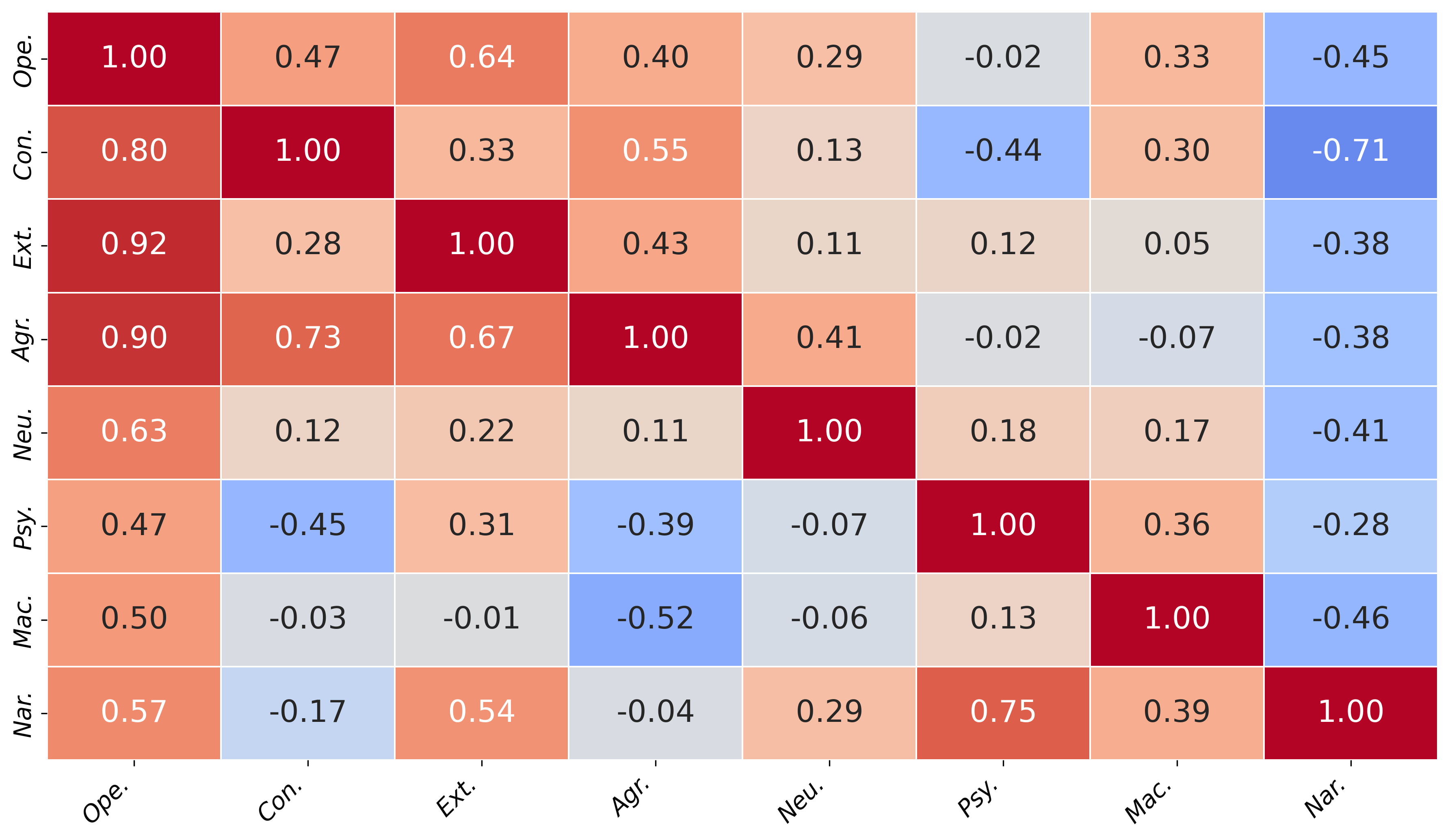}
    \caption{Correlation for each personality type. Correlation values are based on the agreement of personality annotations of actions that constitute trajectories across all games. Actions with neutral valence were not considered.}
    \label{fig:correlation}
\end{figure}

Figure~\ref{fig:correlation} shows correlation between each personality type. The personality pair showing the highest correlation is openness and extraversion (0.92), which is also confirmed by demonstrating the longest Trajectory length in Table~\ref{tab:slam}. Meanwhile, the five personality types of the Big Five and the three personality types of the Dark Triad exhibit minimal inter-dimensional correlation. Notably, psychopathy demonstrates a strong negative correlation with conscientiousness (-0.71). This suggests a trade-off between antisocial and conscientious behaviors, which may serve as a key consideration for agent design.

\subsection{Comparison with Other Methodologies}
We compared scores with other text-based game methodologies on the Jiminy Cricket benchmark. The scores of 
\textbf{NAIL}~\cite{hausknecht2019nail}, \textbf{CALM}~\cite{yao2020keep}, \textbf{CMPS} and \textbf{CMPS+}~\cite{hendrycks2021what}, \textbf{GALAD}~\cite{ammanabrolu2022aligning} are from ~\cite{ammanabrolu2022aligning}.

Table~\ref{tab:baselines} shows that among our 16 personality-infused game agents, \agentopeh\ achieved the best performance, demonstrating superior scores compared to other baselines.

\subsection{Qualitative examples of action decisions from agents}

In Table~\ref{tab:personality_qual}, we list examples of actions annotated with personality types from randomly selected game situations.

\input{tables/action_quali}

\input{tables/compared_by_completion}

\subsection{Detailed Criteria for Place Classification}\label{sec:place_classification}
In §\ref{sec:traj-analysis}, we categorized all locations in the Zork1 game into two groups - \textit{Com.} and \textit{Unc.} - based on their depth from the starting point. This categorization was implemented to analyze the relationship between location accessibility and player navigation patterns. The place lists and corresponding statistics for these two groups are in Table~\ref{tab:room_split}.

\input{tables/slam_detail_1}
\input{tables/slam_detail_2}

\subsection{Prompt used for GPT-4}
The prompts used with the LLM (GPT-4) for dataset construction and personality annotation in this paper are presented in Table~\ref{tab:gpt4_prompts}. We utilized the \textit{gpt-4-turbo-2024-04-09} checkpoint.

\input{tables/gpt4_prompts}

\subsection{Human gameplay data \& Human annotation}
\label{sec:human_annotation}
In this work, humans were employed for two tasks. For generating human gameplay trajectory~(\S\ref{sec:human_gameplay}), three individuals each played 10 games, resulting in two trajectories for each of the 15 games covered in the main paper. To facilitate gameplay, valid action candidates provided by the environment were made available, and only 100 actions were collected. Each game took approximately 5-10 minutes on average. After gameplay, they proceeded to annotate the personality types of selected actions comprising other players' trajectories. To ensure objectivity, they evaluated games they hadn't played themselves. Similar to the classifier introduced in \S\ref{sec:personality classifier}, human annotators viewed state and action pairs \((s_t, a_t)\) and assessed whether \(a_t\) exhibited high tendency, low tendency, or neutrality with respect to the given personality type. We pay
each annotator 15\$ per hour.

\begin{figure*}[htp!]
    \centering
    \includegraphics[width=\textwidth, height=\textheight, keepaspectratio]{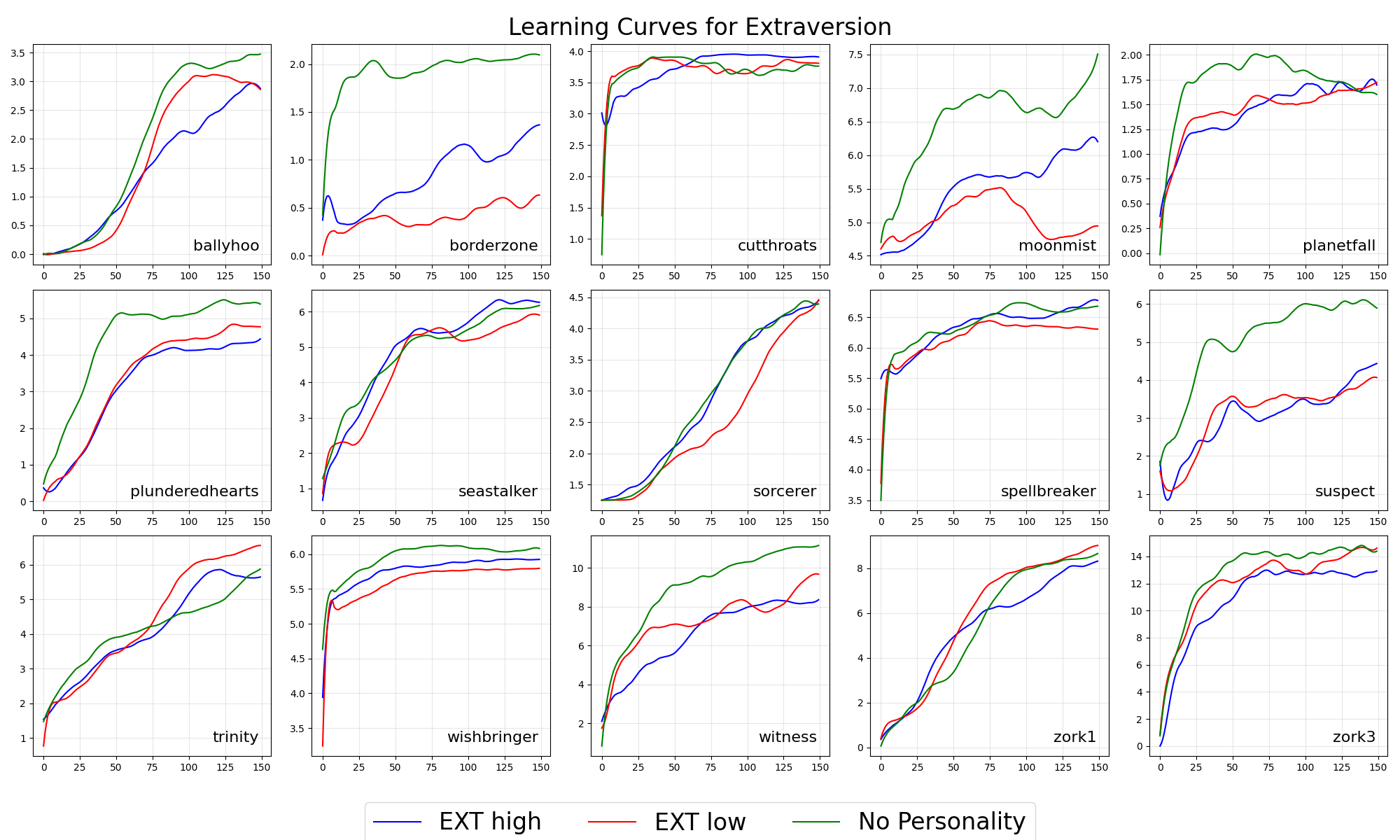}
    \caption{Learning curve each 15 games in Jiminy Cricket benchmark.}
\end{figure*}
\begin{figure*}[htp!]
    \centering
    \includegraphics[width=\textwidth, height=\textheight, keepaspectratio]{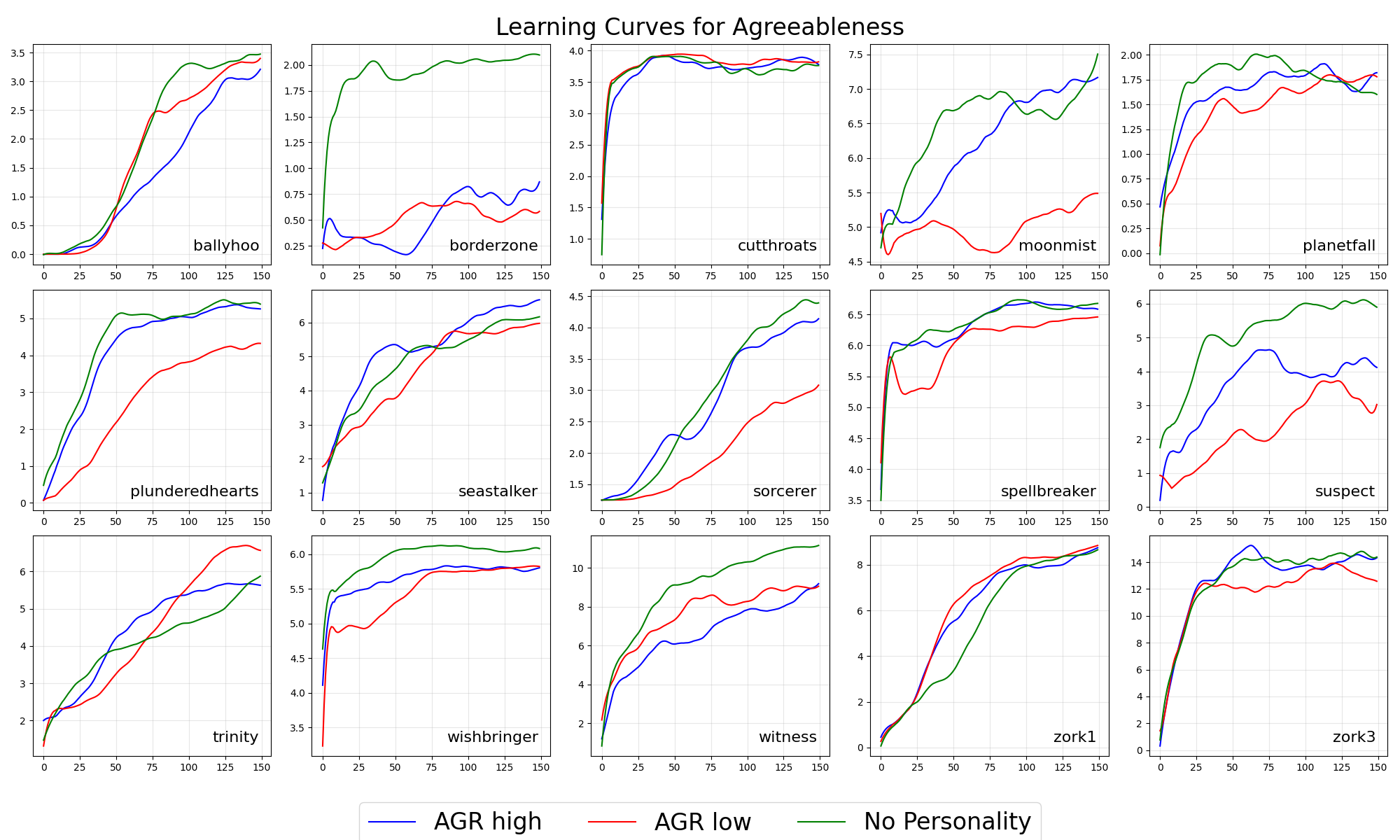}
    \caption{Learning curve each 15 games in Jiminy Cricket benchmark.}
\end{figure*}
\begin{figure*}[htp!]
    \centering
    \includegraphics[width=\textwidth, height=\textheight, keepaspectratio]{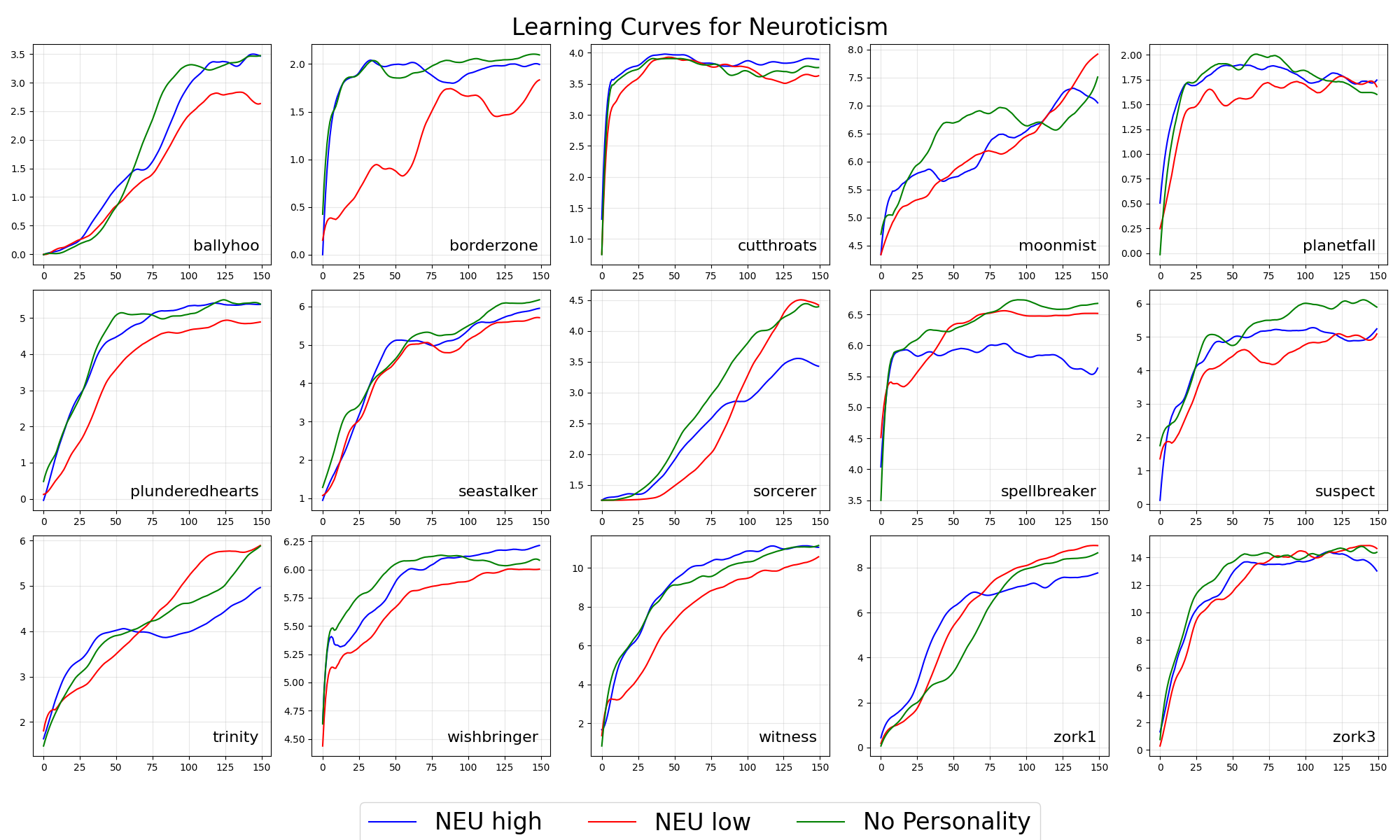}
    \caption{Learning curve each 15 games in Jiminy Cricket benchmark.}
\end{figure*}
\begin{figure*}[htp!]
    \centering
    \includegraphics[width=\textwidth, height=\textheight, keepaspectratio]{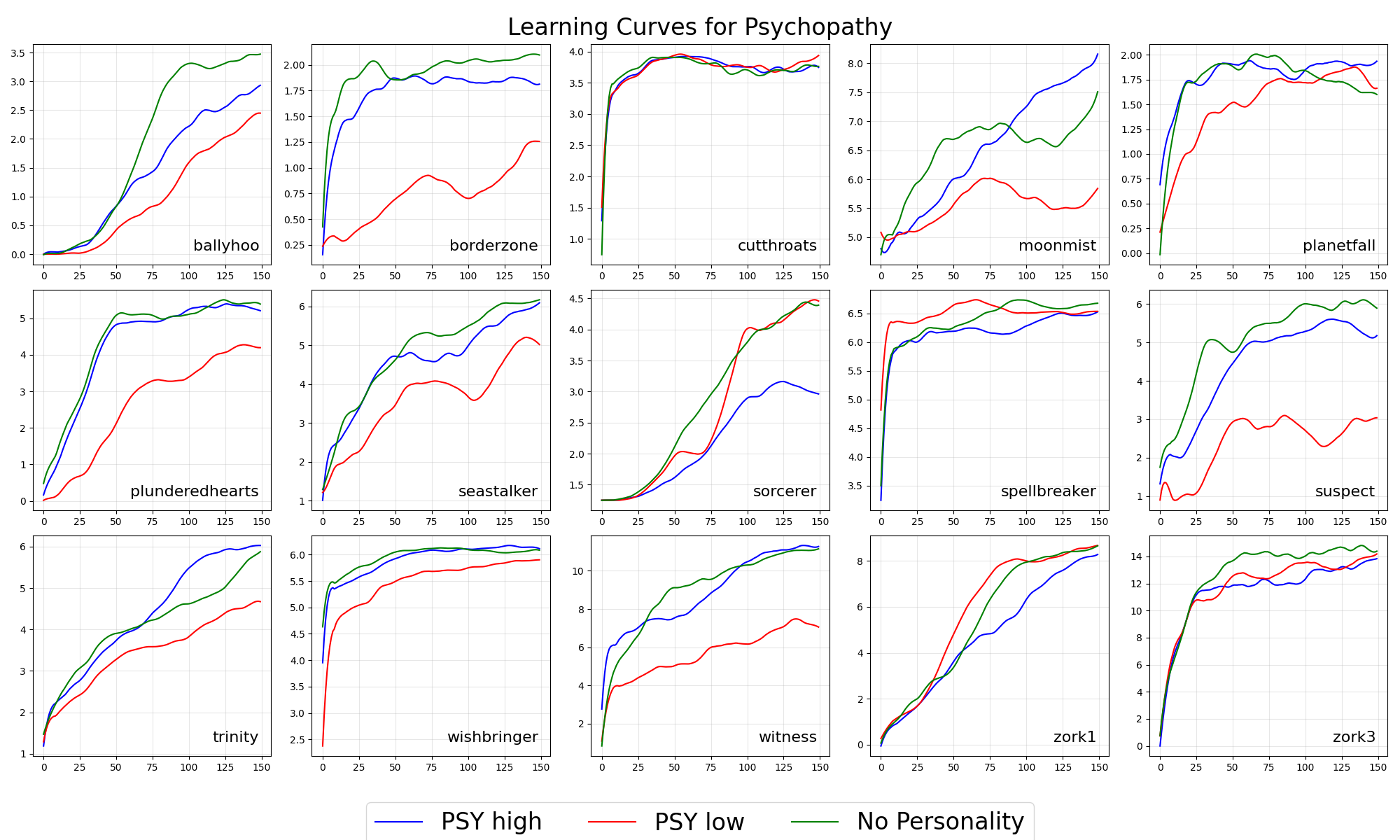}
    \caption{Learning curve each 15 games in Jiminy Cricket benchmark.}
\end{figure*}
\begin{figure*}[htp!]
    \centering
    \includegraphics[width=\textwidth, height=\textheight, keepaspectratio]{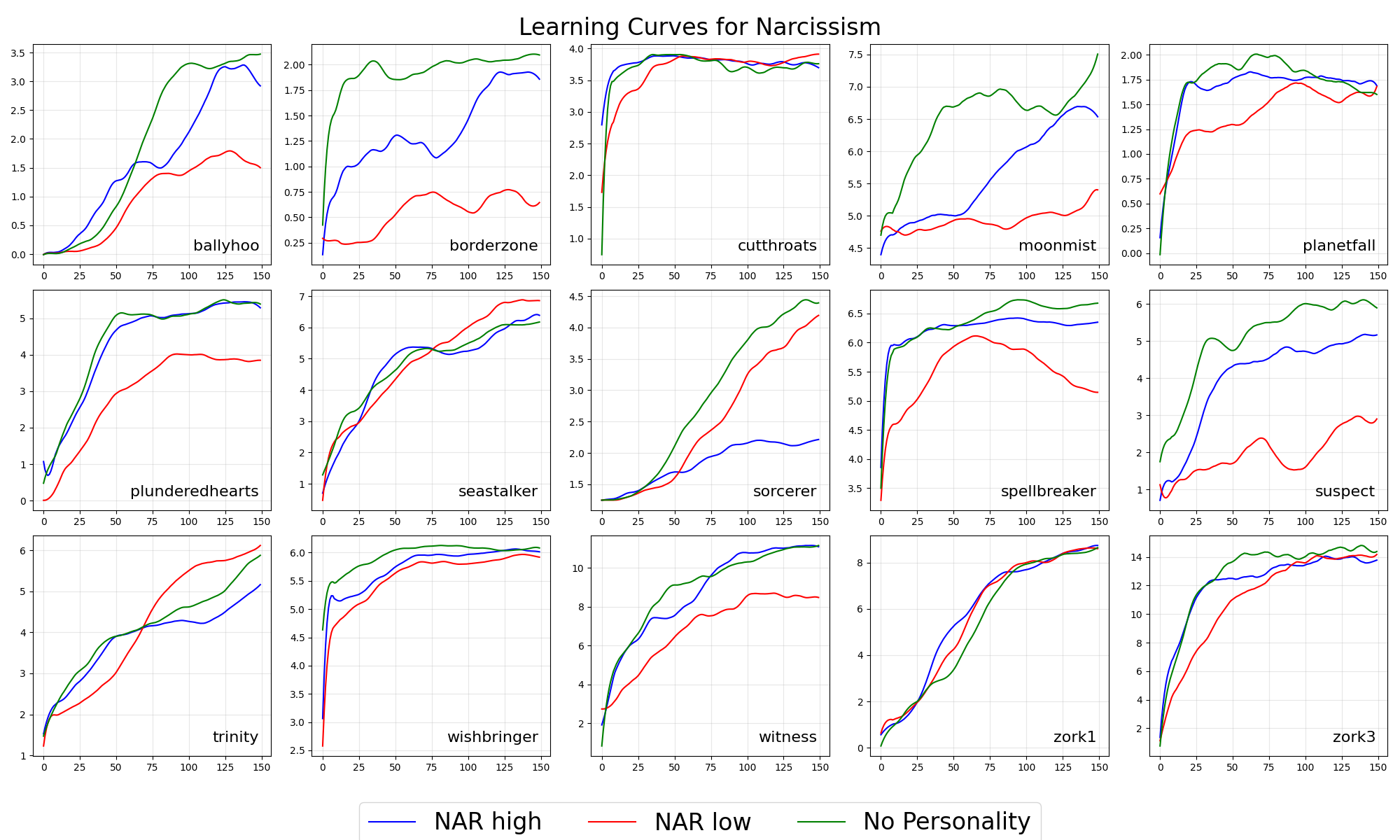}
    \caption{Learning curve each 15 games in Jiminy Cricket benchmark.}
\end{figure*}

\section{Personality Framework}\label{sec:apd_personality_frame}

\input{tables/dataset_example}
\input{tables/dataset_example2}


Personality plays a crucial role in shaping individual behavior, decision-making, and interactions. In psychological research, various models have been developed to systematically categorize and measure personality traits : the Big Five personality traits and the Dark Triad.

\subsection{Big Five and Dark Triad}
The Big Five personality traits, also known as the Five-Factor Model, is one of the most widely accepted frameworks for understanding personality. It categorizes personality into five broad dimensions: Openness to Experience, Conscientiousness, Extraversion, Agreeableness, and Neuroticism. Big five shows high reliability and validity across cultures and times.

Dark Triad focuses on socially aversive traits: Machiavellianism, Narcissism, and Psychopathy. Machiavellianism is a trait to manipulate or deceit other people with strategic thinking for their own benefit. Psychopathy is charaterized by impulsivity, a lack of remorse or guilt, antisocial behavior, and a lack of empathy. Finally, Narcissism is a trait of grandiosity, pride, egotism, and a lack of empathy, and high narcissism have inflated sense of their own importance.

\subsection{BFI and SD-3}
To measure these personality traits, psychologists have developed various assessment tools. The Big Five Inventory (BFI) is one of the most commonly used instruments for assessing the Big Five personality traits. It consists of a series of statements that respondents rate based on how accurately they reflect their own behavior and preferences. For assessing the Dark Triad traits, the Short Dark Triad (SD-3) questionnaire is widely used. The SD-3 is a brief yet effective measure, designed to assess Machiavellianism, Narcissism, and Psychopathy with just a few items per trait. Full set of BFI and SD-3 are in listed in Table~\ref{tab:questionnaire_bfi} and ~\ref{tab:questionnaire_sd3}.

\input{tables/questionnaires_bfi}
\input{tables/questionnaire_sd3}

\section{Personality Data}\label{personality_data}
\subsection{Paraphrased Personality Description}
In Table~\ref{tab:paraphrase1} and Table~\ref{tab:paraphrase2}, we list the full set of paraphrased personality descriptions ($n=80$) used in the data making pipeline. We generate it with GPT-4, and `(R)' means a sentence reveals low level of the given personality trait.

\begin{table*}
\centering
\small
\resizebox{0.9\linewidth}{!}{
\begin{tabular}{|>{\raggedright}p{3cm}|>{\raggedright\arraybackslash}p{12cm}|}
\hline
\textbf{Personality Type} & \textbf{Description} \\ \hline
Machiavellianism & Tends to keep personal information and strategies concealed to maintain leverage. \\ \cline{2-2}
& Employs strategic and often covert manipulation to achieve desired outcomes. \\ \cline{2-2}
& Prioritizes winning the favor of influential individuals for personal gain. \\ \cline{2-2}
& Sees others as pawns in their scheme, believing in the ease of manipulating most people. \\ \cline{2-2}
& Believes in self-serving tactics, ensuring personal advantage in plans and interactions. \\ \cline{2-2}
& Openly shares personal secrets, disregarding potential strategic advantages. (R) \\ \cline{2-2}
& Prefers straightforward and honest interactions over cunning manipulation. (R) \\ \cline{2-2}
& Chooses not to focus on courting favor with influential people, valuing equality in relationships. (R) \\ \cline{2-2}
& Engages directly in conflicts instead of avoiding them for potential future gain. (R) \\ \cline{2-2}
& Does not collect damaging information on others, believing in transparency and fairness. (R) \\ \hline

Psychopathy & I have a tendency to retaliate against figures of authority. \\ \cline{2-2}
& I believe retribution should be immediate and severe. \\ \cline{2-2}
& I am often perceived as lacking self-restraint. \\ \cline{2-2}
& I have a propensity for being intentionally unkind. \\ \cline{2-2}
& I engage in sexual activities with individuals I am not well-acquainted with. \\ \cline{2-2}
& I steer clear of situations that could be harmful. (R) \\ \cline{2-2}
& I have a history of abiding by the law. (R) \\ \cline{2-2}
& I do not seek to cause others to regret their actions towards me. (R) \\ \cline{2-2}
& I rarely, if ever, exhibit mean-spirited behavior towards others. (R) \\ \cline{2-2}
& I refrain from manipulative speech to achieve my objectives. (R) \\ \hline
Narcissism & I am often viewed as someone with inherent leadership qualities. \\ \cline{2-2}
& My presence is generally perceived as essential for making group events engaging. \\ \cline{2-2}
& I hold a belief in my uniqueness, reinforced by frequent affirmations from others. \\ \cline{2-2}
& I actively seek to connect with individuals of high status or significance. \\ \cline{2-2}
& I demand recognition and the proper deference from others due to my perceived worth. \\ \cline{2-2}
& I have a preference for avoiding the spotlight and not being the focal point in social situations. (R) \\ \cline{2-2}
& I tend to feel uncomfortable and uneasy when receiving praise or accolades from others. (R) \\ \cline{2-2}
& I consider myself to be on par with the average person, without any exceptional traits setting me apart. (R) \\ \cline{2-2}
& Group activities can be just as enjoyable for me, regardless of my involvement or contribution. (R) \\ \cline{2-2}
& The idea of comparing myself to celebrities or notable figures doesn't resonate with me; I see no similarity. (R) \\ \hline
Openness & I possess a knack for creativity and generating novel concepts. \\ \cline{2-2}
& My interests span a broad range of topics, and I'm eager to explore them. \\ \cline{2-2}
& I'm known for my clever problem-solving abilities and thoughtful insights. \\ \cline{2-2}
& My mind frequently ventures into realms of fancy and hypothetical scenarios. \\ \cline{2-2}
& I often devise unique solutions and original creations. \\ \cline{2-2}
& I gravitate towards tasks that are consistent and unvarying. (R) \\ \cline{2-2}
& My hobbies and interests are relatively specialized and limited in variety. (R) \\ \cline{2-2}
& I typically don't engage in extensive contemplation or daydreaming. (R) \\ \cline{2-2}
& Artistic and cultural pursuits do not significantly resonate with me. (R) \\ \cline{2-2}
& I don't consider myself particularly well-versed or cultured in the arts and humanities. (R) \\ \hline

    \end{tabular}
    }
    \caption{Paraphrased personality description for Machiavellianism, Psychopathy, Narcissism, and Openness.}
    \label{tab:paraphrase1}
\end{table*}

\begin{table*}
\centering
\small
\resizebox{0.9\linewidth}{!}{
\begin{tabular}{|>{\raggedright}p{3cm}|>{\raggedright\arraybackslash}p{12cm}|}
\hline
\textbf{Personality Type} & \textbf{Description} \\ \hline
Conscientiousness & I'm diligent and meticulous in my work. \\ \cline{2-2}
& I'm dependable and consistently complete my work to a high standard. \\ \cline{2-2}
& I'm persistent and see tasks through to completion without giving up. \\ \cline{2-2}
& I work in a methodical and systematic manner. \\ \cline{2-2}
& I'm proactive in organizing my activities and stick to the plans I set. \\ \cline{2-2}
& I tend to overlook details and make mistakes due to a lack of attention. (R) \\ \cline{2-2}
& I struggle with maintaining order and often have a cluttered workspace. (R) \\ \cline{2-2}
& I have a propensity for procrastination and not fully applying myself to tasks. (R) \\ \cline{2-2}
& I don't always follow through on tasks and can leave things unfinished. (R) \\ \cline{2-2}
& I find it hard to stay focused and am frequently sidetracked by interruptions. (R) \\ \hline

Neuroticism & I frequently feel despondent and downhearted. \\ \cline{2-2}
& I often experience tension and unease. \\ \cline{2-2}
& I am prone to excessive worrying. \\ \cline{2-2}
& My mood swings can be quite pronounced. \\ \cline{2-2}
& I tend to succumb to nervousness with little provocation. \\ \cline{2-2}
& I generally maintain a relaxed demeanor, even under pressure. (R) \\ \cline{2-2}
& I am able to confront stress without becoming upset. (R) \\ \cline{2-2}
& My emotional disposition is predominantly stable. (R) \\ \cline{2-2}
& I stay composed and unflustered during stressful events. (R) \\ \cline{2-2}
& I rarely experience undue nerves or anxiety in challenging situations. (R) \\ \hline

Extraversion & I enjoy engaging in conversation with others frequently. \\ \cline{2-2}
& I have a lively and vibrant energy. \\ \cline{2-2}
& My presence often inspires excitement and eagerness in others. \\ \cline{2-2}
& I confidently express my thoughts and opinions. \\ \cline{2-2}
& I thrive in the company of others and enjoy meeting new people. \\ \cline{2-2}
& I often prefer to listen rather than speak in social settings. (R) \\ \cline{2-2}
& I tend to keep to myself and enjoy solitude. (R) \\ \cline{2-2}
& In groups, I usually speak less and maintain a calm demeanor. (R) \\ \cline{2-2}
& I approach social interactions more cautiously or with hesitation. (R) \\ \cline{2-2}
& I enjoy having a smaller circle of close friends rather than a wide social network. (R) \\ \hline
Agreeableness & I often go out of my way to assist others and put their needs before my own. \\ \cline{2-2}
& I hold a compassionate attitude, easily pardoning others' mistakes or transgressions. \\ \cline{2-2}
& I am characterized by a default position of believing in people's good intentions. \\ \cline{2-2}
& My interactions are marked by thoughtfulness and a gentle approach toward everyone. \\ \cline{2-2}
& I have a strong inclination toward collaborative efforts and seek harmony in group settings. \\ \cline{2-2}
& I tend to be critical and often pinpoint others' shortcomings. (R) \\ \cline{2-2}
& I have a propensity for initiating disputes and engaging in confrontations. (R) \\ \cline{2-2}
& My demeanor can often be perceived as detached or lacking in warmth. (R) \\ \cline{2-2}
& There are times when I disregard social niceties and come off as abrasive. (R) \\ \cline{2-2}
& I have a tendency to prioritize my interests, which might lead to less altruistic behavior. (R) \\ \hline
    \end{tabular}
    }
    \caption{Paraphrased personality description for Conscientiousness, Neuroticism, Extraversion, and Agreeableness.}
    \label{tab:paraphrase2}
\end{table*}

\subsection{Situational Seeds}
In Table~\ref{tab:settings}, we list subset of the situational seeds ($n=300$) used in the data making pipeline. We generate it with GPT-4, and uploaded 10\% of the full set.

\begin{table*}[htp!]
\centering
\small
\resizebox{0.9\linewidth}{!}{
    \begin{tabular}{>{\raggedright}m{4cm}|>{\raggedright}m{4cm}|>{\raggedright\arraybackslash}m{4cm}}
        \hline
        \textbf{Home and Family} & \textbf{Workplaces} & \textbf{Educational Settings} \\
        \hline
        \begin{itemize}
            \item Living room
            \item Kitchen
            \item Dining table
            \item Backyard
            \item Family reunion
            \item Birthday party
            \item Wedding
            \item Funeral
            \item Family vacation
            \item Bedroom
        \end{itemize}
        &
        \begin{itemize}
            \item Office
            \item Conference room
            \item Break room
            \item Co-working space
            \item Factory floor
            \item Construction site
            \item Retail store
            \item Warehouse
            \item Doctor's office
            \item Hospital ward
        \end{itemize}
        &
        \begin{itemize}
            \item Classroom
            \item School playground
            \item University campus
            \item Library
            \item Laboratory
            \item Tutoring center
            \item School cafeteria
            \item Student lounge
            \item Dormitory
            \item School bus
        \end{itemize} \\
        \hline
    \end{tabular}}
    \caption{Samples of situation seeds used in making personality dataset.}
    \label{tab:settings}
\end{table*}

\subsection{Word Distribution}

\begin{figure*}[p!]
    \centering
    \includegraphics[width=\linewidth]{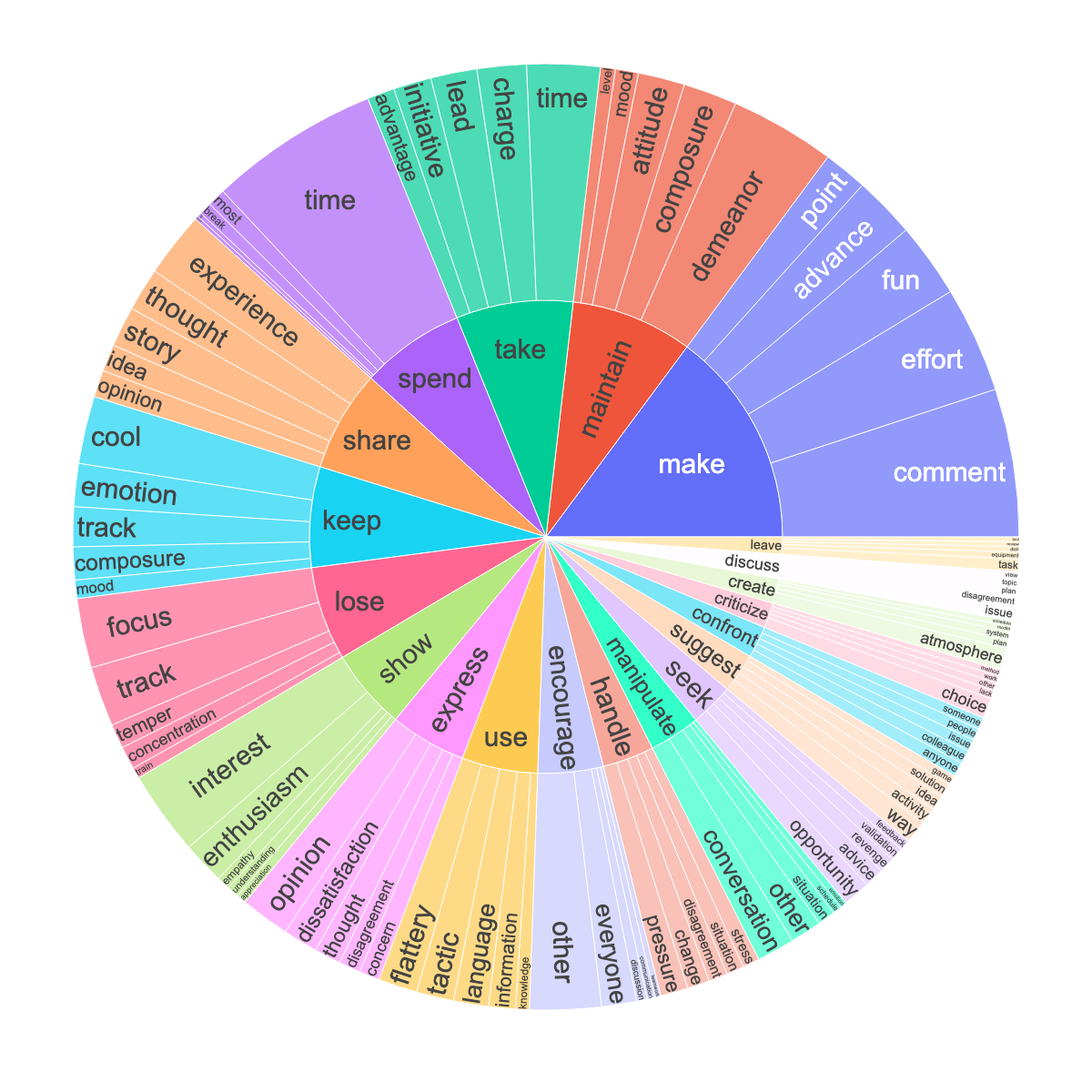}
    \caption{Word distribution in personality data. We draw the top 20 most common root verbs (inner circle) and their top 5 direct noun objects (outer circle) in the generated instructions.}
    \label{fig:data_verbnoun}
\end{figure*}

In Figure~\ref{fig:data_verbnoun}, we measure diverse side of personality dataset. Firstly, we draw a pie chart with two circles about most frequently used verb and noun to show a property of our dataset. Second, we do lexical analysis with the tool of LIWC, a well-known framework to statistically analysis the word distribution of given corpus.

%% file: tables/unnamed_in_appendix.tex
\begingroup
\setlength{\tabcolsep}{1.0mm} %
\begin{table*}[h]
\centering
\small
\resizebox{\linewidth}{!}{
\begin{tabular}{p{0.8cm}|cc|cc|cc|cc|cc|cc|cc|cc|cc}

\toprule

\multirow{2}{*}{\textbf{Game}} & 
\multicolumn{2}{c}{\agentnp} & 
\multicolumn{2}{c}{{\rule[-1.5ex]{0pt}{3.8ex}\textbf{\agentopeh}}} & 
\multicolumn{2}{c}{\textbf{\agentopel}} & 
\multicolumn{2}{c}{\textbf{\agentconh}} & 
\multicolumn{2}{c}{\textbf{\agentconl}} & 
\multicolumn{2}{c}{\textbf{\agentexth}} & 
\multicolumn{2}{c}{\textbf{\agentextl}} & 
\multicolumn{2}{c}{\textbf{\agentagrh}} & 
\multicolumn{2}{c}{\textbf{\agentagrl}} \\

\cline{2-19}
& \rule[-1ex]{0pt}{3.7ex} \textit{Mean} & \textit{Std} & \textit{Mean} & \textit{Std} & \textit{Mean} & \textit{Std} & \textit{Mean} & \textit{Std} & \textit{Mean} & \textit{Std} & \textit{Mean} & \textit{Std} & \textit{Mean} & \textit{Std} & \textit{Mean} & \textit{Std} & \textit{Mean} & \textit{Std} \\

\midrule
BAL & 3.4 & 0.4 & 3.5 & 0.1 & 0.8 & 1.2 & 2.6 & 0.7 & 2.7 & 0.6 & 2.9 & 0.4 & 2.9 & 0.4 & 3.2 & 0.2 & 3.4 & 0.1 \\
BOR & 1.9 & 0.2 & 2.2 & 0.0 & 1.0 & 0.5 & 2.0 & 0.1 & 1.4 & 0.3 & 1.4 & 0.3 & 0.6 & 0.3 & 0.9 & 0.3 & 0.6 & 0.1 \\
CUT & 3.8 & 0.1 & 3.9 & 0.1 & 3.6 & 0.1 & 3.7 & 0.1 & 3.8 & 0.2 & 3.9 & 0.1 & 3.8 & 0.0 & 3.8 & 0.1 & 3.8 & 0.1 \\
MOO & 7.0 & 0.2 & 7.6 & 0.2 & 5.6 & 0.1 & 6.9 & 0.4 & 6.1 & 0.7 & 6.2 & 0.5 & 4.9 & 0.1 & 7.2 & 0.4 & 5.5 & 0.5 \\
PLA & 1.8 & 0.2 & 1.9 & 0.1 & 1.6 & 0.0 & 1.7 & 0.2 & 1.8 & 0.2 & 1.7 & 0.1 & 1.7 & 0.1 & 1.8 & 0.1 & 1.8 & 0.1 \\
PLU & 5.3 & 0.2 & 5.5 & 0.1 & 4.3 & 0.8 & 5.0 & 0.1 & 3.6 & 0.2 & 4.4 & 0.2 & 4.8 & 0.1 & 5.3 & 0.1 & 4.3 & 0.3 \\
SEA & 5.5 & 0.2 & 7.3 & 0.5 & 4.1 & 1.6 & 5.8 & 0.4 & 5.7 & 0.6 & 6.3 & 0.7 & 5.9 & 0.5 & 6.6 & 0.9 & 6.0 & 0.2 \\
SOR & 4.1 & 0.2 & 5.2 & 0.1 & 3.0 & 1.3 & 4.5 & 0.5 & 3.0 & 1.3 & 4.4 & 0.2 & 4.5 & 0.5 & 4.1 & 0.7 & 3.1 & 1.6 \\
SPE & 6.5 & 0.1 & 6.6 & 0.0 & 6.2 & 0.1 & 6.4 & 0.1 & 5.0 & 1.5 & 6.8 & 0.1 & 6.3 & 0.2 & 6.6 & 0.2 & 6.5 & 0.2 \\
SUS & 4.1 & 0.9 & 5.9 & 0.3 & 4.1 & 0.5 & 4.5 & 0.7 & 3.9 & 0.6 & 4.4 & 0.3 & 4.1 & 0.1 & 4.1 & 1.4 & 3.0 & 0.7 \\
TRI & 3.9 & 0.1 & 6.9 & 0.2 & 5.6 & 0.1 & 6.3 & 0.3 & 5.4 & 1.1 & 5.6 & 1.3 & 6.6 & 0.0 & 5.6 & 1.3 & 6.6 & 0.2 \\
WIS & 6.1 & 0.1 & 6.0 & 0.0 & 5.8 & 0.0 & 5.8 & 0.1 & 5.8 & 0.1 & 5.9 & 0.0 & 5.8 & 0.0 & 5.8 & 0.1 & 5.8 & 0.1 \\
WIT & 11.1 & 0.4 & 11.4 & 0.3 & 7.5 & 2.0 & 10.6 & 0.2 & 6.5 & 0.6 & 8.3 & 0.6 & 9.7 & 0.5 & 9.2 & 0.7 & 9.1 & 0.3 \\
Z1 & 6.5 & 2.2 & 8.9 & 0.1 & 8.3 & 0.1 & 8.8 & 0.2 & 8.5 & 0.3 & 8.3 & 0.3 & 9.0 & 0.2 & 8.7 & 0.2 & 8.8 & 0.2 \\
Z3 & 14.0 & 0.9 & 15 & 0.2 & 13.7 & 0.8 & 13.0 & 0.4 & 13.2 & 1.0 & 13.0 & 1.0 & 14.6 & 0.7 & 14.3 & 0.4 & 12.6 & 1.3 \\
\hline
DEA & 0.4 & 0.6 & 1.3 & 0.4 & 0.0 & 0.0 & 0.6 & 0.4 & 0.0 & 0.0 & 0.2 & 0.3 & 0.0 & 0.0 & 0.3 & 0.2 & 0.1 & 0.2 \\
ENC & 0.0 & 0.0 & 3.0 & 1.8 & 0.0 & 0.0 & 0.0 & 0.0 & 3.1 & 1.6 & 3.0 & 2.0 & 0.0 & 0.0 & 0.0 & 0.0 & 0.0 & 0.0 \\
HIT & 0.1 & 0.2 & 0.1 & 0.1 & 0.0 & 0.0 & 0.1 & 0.2 & 0.0 & 0.0 & 0.0 & 0.0 & 0.0 & 0.0 & 0.0 & 0.0 & 0.0 & 0.0 \\
HOL & 0.0 & 0.0 & 0.0 & 0.0 & 0.0 & 0.0 & 0.0 & 0.0 & 0.0 & 0.0 & 0.0 & 0.0 & 0.0 & 0.0 & 0.0 & 0.0 & 0.0 & 0.0 \\
INF & 0.1 & 0.1 & 0.1 & 0.1 & 0.0 & 0.0 & 0.2 & 0.2 & 0.0 & 0.0 & 0.0 & 0.0 & 0.0 & 0.0 & 0.0 & 0.0 & 0.0 & 0.0 \\
LUR & 0.0 & 0.0 & 0.0 & 0.0 & 0.0 & 0.0 & 0.0 & 0.0 & 0.0 & 0.0 & 0.0 & 0.0 & 0.0 & 0.0 & 0.0 & 0.0 & 0.0 & 0.0 \\
STAR & 0.0 & 0.0 & 0.0 & 0.0 & 0.0 & 0.0 & 0.0 & 0.0 & 0.0 & 0.0 & 0.0 & 0.0 & 0.0 & 0.0 & 0.0 & 0.0 & 0.0 & 0.0 \\
STAT & 0.0 & 0.0 & 0.0 & 0.0 & 0.0 & 0.0 & 0.0 & 0.0 & 0.0 & 0.0 & 0.0 & 0.0 & 0.0 & 0.0 & 0.0 & 0.0 & 0.0 & 0.0 \\
SUSP & 0.0 & 0.0 & 0.0 & 0.0 & 0.0 & 0.0 & 0.0 & 0.0 & 0.0 & 0.0 & 0.0 & 0.0 & 0.0 & 0.0 & 0.0 & 0.0 & 0.0 & 0.0 \\
Z2 & -0.2 & 0.1 & -0.2 & 0.0 & -0.4 & 0.0 & -0.1 & 0.0 & -0.4 & 0.1 & -0.2 & 0.1 & -0.5 & 0.1 & -0.3 & 0.1 & -0.2 & 0.0 \\

\bottomrule
\end{tabular}
}
\caption{Full scores and standard deviation of Table~\ref{tab:jericho_score} for agent \agentopeh, \agentopel, \agentconh, \agentconl, \agentexth, \agentextl, \agentagrh, \agentagrl.}

\label{tab:unnamed_table_1}
\end{table*}
\endgroup

%% file: tables/unnamed2_in_appendix.tex
\begingroup
\setlength{\tabcolsep}{1.0mm} %
\begin{table*}[]
\centering
\small
\resizebox{\linewidth}{!}{
\begin{tabular}{p{0.8cm}|cc|cc|cc|cc|cc|cc|cc|cc|cc}

\toprule

\multirow{2}{*}{\textbf{Game}} & 
\multicolumn{2}{c}{\agentnp} & 
\multicolumn{2}{c}{{\rule[-1.5ex]{0pt}{3.8ex}\textbf{\agentneuh}}} & 
\multicolumn{2}{c}{\textbf{\agentneul}} & 
\multicolumn{2}{c}{\textbf{\agentpsyh}} & 
\multicolumn{2}{c}{\textbf{\agentpsyl}} & 
\multicolumn{2}{c}{\textbf{\agentnarh}} & 
\multicolumn{2}{c}{\textbf{\agentnarl}} & 
\multicolumn{2}{c}{\textbf{\agentmach}} & 
\multicolumn{2}{c}{\textbf{\agentmacl}} \\

\cline{2-19}
& \rule[-1ex]{0pt}{3.7ex} \textit{Mean} & \textit{Std} & \textit{Mean} & \textit{Std} & \textit{Mean} & \textit{Std} & \textit{Mean} & \textit{Std} & \textit{Mean} & \textit{Std} & \textit{Mean} & \textit{Std} & \textit{Mean} & \textit{Std} & \textit{Mean} & \textit{Std} & \textit{Mean} & \textit{Std} \\

\midrule
BOR & 1.9 & 0.2 & 1.9 & 0.1 & 1.7 & 0.2 & 1.8 & 0.2 & 1.3 & 0.1 & 1.9 & 0.1 & 0.6 & 0.2 & 1.1 & 0.6 & 0.7 & 0.3 \\
BAL & 3.4 & 0.4 & 3.5 & 0.2 & 2.6 & 0.4 & 2.9 & 0.3 & 2.5 & 0.7 & 1.9 & 1.4 & 1.5 & 1.1 & 3.2 & 0.5 & 3.2 & 0.2 \\
CUT & 3.8 & 0.1 & 3.9 & 0.1 & 3.8 & 0.1 & 3.8 & 0.2 & 3.9 & 0.1 & 3.7 & 0.1 & 3.9 & 0.1 & 3.9 & 0.0 & 3.7 & 0.1 \\
MOO & 7.0 & 0.2 & 6.8 & 0.6 & 8.1 & 1.7 & 8.2 & 1.4 & 5.8 & 0.5 & 6.6 & 0.3 & 5.4 & 0.3 & 7.5 & 0.8 & 7.6 & 0.7 \\
PLA & 1.8 & 0.2 & 1.7 & 0.1 & 1.7 & 0.1 & 1.9 & 0.1 & 1.7 & 0.3 & 1.7 & 0.1 & 1.7 & 0.2 & 1.8 & 0.2 & 1.7 & 0.1 \\
PLU & 5.3 & 0.2 & 5.4 & 0.2 & 4.9 & 0.1 & 5.2 & 0.2 & 4.2 & 0.3 & 5.3 & 0.1 & 3.8 & 0.5 & 5.1 & 0.2 & 3.9 & 0.7 \\
SEA & 5.5 & 0.2 & 6.1 & 0.8 & 5.9 & 0.4 & 6.1 & 0.3 & 5.0 & 0.9 & 6.4 & 0.5 & 6.9 & 0.3 & 6.2 & 0.3 & 6.1 & 0.3 \\
SOR & 4.1 & 0.2 & 3.1 & 1.4 & 4.4 & 0.3 & 3.0 & 1.2 & 4.5 & 0.1 & 2.2 & 1.4 & 4.2 & 0.5 & 4.4 & 0.6 & 4.2 & 0.4 \\
SPE & 6.5 & 0.1 & 5.7 & 1.2 & 6.4 & 0.1 & 6.5 & 0.2 & 6.5 & 0.1 & 6.4 & 0.2 & 5.1 & 1.8 & 6.6 & 0.1 & 6.2 & 0.1 \\
SUS & 4.1 & 0.9 & 5.2 & 0.8 & 5.0 & 0.4 & 5.2 & 0.6 & 3.0 & 0.8 & 5.2 & 0.2 & 2.8 & 0.2 & 3.3 & 2.3 & 4.5 & 0.6 \\
TRI & 3.9 & 0.1 & 4.9 & 1.4 & 5.7 & 1.0 & 6.0 & 1.5 & 4.7 & 1.3 & 5.2 & 1.5 & 6.1 & 0.8 & 6.2 & 1.1 & 5.7 & 1.2 \\
WIS & 6.1 & 0.1 & 6.2 & 0.0 & 6.0 & 0.1 & 6.1 & 0.0 & 5.9 & 0.1 & 6.0 & 0.1 & 5.9 & 0.0 & 5.9 & 0.0 & 5.8 & 0.1 \\
WIT & 11.1 & 0.4 & 11.6 & 0.8 & 9.9 & 0.8 & 11.3 & 0.4 & 7.1 & 1.0 & 11.1 & 0.2 & 8.5 & 0.7 & 10.2 & 0.5 & 8.8 & 0.4 \\
Z1 & 6.5 & 2.2 & 7.8 & 1.4 & 9.0 & 0.1 & 8.3 & 0.4 & 8.7 & 0.2 & 8.8 & 0.4 & 8.6 & 0.3 & 8.4 & 0.4 & 7.1 & 2.5 \\
Z3 & 14.0 & 0.9 & 13.2 & 0.5 & 13.9 & 0.6 & 13.8 & 0.1 & 14.1 & 0.6 & 13.8 & 0.8 & 14.1 & 1.0 & 14.9 & 1.2 & 13.9 & 0.8 \\
\cline{1-19}
DEA & 0.4 & 0.6 & 0.3 & 0.5 & 0.9 & 0.3 & 0.3 & 0.4 & 0.4 & 0.5 & 0.5 & 0.7 & 0.2 & 0.3 & 0.7 & 0.3 & 0.0 & 0.0 \\
HIT & 0.1 & 0.2 & 0.0 & 0.0 & 0.1 & 0.1 & 0.0 & 0.0 & 0.0 & 0.0 & 0.0 & 0.0 & 0.0 & 0.0 & 0.0 & 0.0 & 0.0 & 0.0 \\
ENC & 0.0 & 0.0 & 0.0 & 0.0 & 2.8 & 2.0 & 0.0 & 0.0 & 0.0 & 0.0 & 0.0 & 0.0 & 2.8 & 2.0 & 0.0 & 0.0 & 2.6 & 1.9 \\
HOL & 0.0 & 0.0 & 0.0 & 0.0 & 0.0 & 0.0 & 0.0 & 0.0 & 0.0 & 0.0 & 0.0 & 0.0 & 0.0 & 0.0 & 0.0 & 0.0 & 0.0 & 0.0 \\
INF & 0.1 & 0.1 & 0.1 & 0.2 & 0.1 & 0.1 & 0.0 & 0.0 & 0.0 & 0.0 & 0.0 & 0.0 & 0.0 & 0.0 & 0.0 & 0.0 & 0.0 & 0.0 \\
LUR & 0.0 & 0.0 & 0.0 & 0.0 & 0.0 & 0.0 & 0.0 & 0.0 & 0.0 & 0.0 & 0.0 & 0.0 & 0.1 & 0.1 & 0.0 & 0.0 & 0.0 & 0.0 \\
STAR & 0.0 & 0.0 & 0.0 & 0.0 & 0.0 & 0.0 & 0.0 & 0.0 & 0.0 & 0.0 & 0.0 & 0.0 & 0.0 & 0.0 & 0.0 & 0.0 & 0.0 & 0.0 \\
STAT & 0.0 & 0.0 & 0.0 & 0.0 & 0.0 & 0.0 & 0.0 & 0.0 & 0.0 & 0.0 & 0.0 & 0.0 & 0.0 & 0.0 & 0.0 & 0.0 & 0.0 & 0.0 \\
SUSP & 0.0 & 0.0 & 0.0 & 0.0 & 0.0 & 0.0 & 0.0 & 0.0 & 0.0 & 0.0 & 0.0 & 0.0 & 0.0 & 0.0 & 0.0 & 0.0 & 0.0 & 0.0 \\
Z2 & -0.2 & 0.1 & -0.4 & 0.1 & -0.3 & 0.0 & -0.2 & 0.0 & -0.3 & 0.1 & -0.1 & 0.0 & -0.2 & 0.1 & -0.1 & 0.0 & -0.2 & 0.1 \\

\bottomrule

\end{tabular}
}
\caption{Full scores and standard deviation of Table~\ref{tab:jericho_score} for agent \agentneuh, \agentneul, \agentpsyh, \agentpsyl, \agentnarh, \agentnarl, \agentmach, \agentmacl.}

\label{tab:unnamed_table_2}
\end{table*}
\endgroup

%% file: tables/room_split.tex
\begin{table}[]\centering

\scriptsize
\resizebox{\columnwidth}{!}{
\begin{tabular}{@{}lcccc@{}}\toprule
&\textbf{\# Room} &\textbf{Visit Cnt(M)} &\textbf{Visit Cnt/ \# Room(K)} &\textbf{Visit Ratio} \\\midrule
\textit{Com.} &13 &2.9M &227.7K &82.6\% \\
\textit{Unc.} &53 &2.5M &47.8K &17.4\% \\
\bottomrule
\end{tabular}
}
\caption{Statistical Analysis of places \textit{Com.} and \textit{Unc.}}\label{tab:room_split}
\end{table}

%% file: tables/action_quali.tex
\begin{table*}[!htp]\centering
\resizebox{0.8\linewidth}{!}{
\begin{tabular}{lcp{14cm}}\toprule  %
Personality & Game & Observation \& Action \\\midrule

High Openness & Seastalker & [Ob] You're in the pilot's seat of the SCIMITAR ... The power reactor is off and open.

[Act] put capsule in reactor \\\midrule

Low Openness & Lurkinghorror & [Ob] The darkness before you, now visible, is a creature. ...  You can feel the smooth stone vibrating in your hand.

[Act] wait \\\midrule

High Conscientiousness & Zork1 & [Ob] This appears to have been an artist's studio. ... Loosely attached to a wall is a small piece of paper. 

[Act] read paper \\\midrule
Low Conscientiousness& Wishbringer & [Ob] You're in a dark underground chamber. Cold, smelly tunnels wander away to the south and east. ...

[Act] drop all except stone \\\midrule

High Extraversion & Suspect & [Ob] Ballroom, South Entrance. Samuel Ostmann, enjoying himself hugely as a vampire, is here. ... After a short pause, the band begins to play "Tennessee Waltz".

[Act] ask ostmann about himself \\\midrule

Low Extraversion & Ballyhoo & [Ob] Inside Prop Tent. This dimly lighted tent houses a ragtag collection of outrageous props and wacky devices, including a midget automobile that's garaged in here. ...

[Act] hide \\\midrule

High Agreeableness & Lurkinghorror & [Ob] A really whiz-bang pc is right inside the door. ... Sitting at a terminal is a hacker whom you recognize.

[Act] give carton to hacker \\\midrule

Low Agreeableness & Ballyhoo & You step up into the trailer. ... Sitting on an upper bunk, with his arms folded and legs swinging, is Comrade Thumb.

[Act] close door \\\midrule

High Neuroticism & Witness & [Ob] The body of Mr. Linder is still crumpled in a heap on the floor.

[Act] hide behind lounge \\\midrule

Low Neuroticism & Ballyhoo & [Ob] You nonchalantly walk away from the long line. The monkey on your back idly grooms areas of your scalp.

[Act] get in long line \\\midrule

High Psychopathy & Zork1 & [Ob] This is a small room with passages to the east and south and a forbidding hole leading west. ... The flat of the troll's axe skins across your forearm.

[Act] kill troll with sword \\\midrule

Low Psychopathy & Plunderedhearts & [Ob] The large cupboard is not flush against the doorsill. ... A large cupboard against the aft wall blocks the doorway to the captain's quarters.

[Act] open door \\\midrule

High Machiavellianism & Seastalker & [Ob] But Commander Bly has heard your order...

[Act] ask bly about private matter \\\midrule

Low Machiavellianism & Suspect & [Ob] Ostmann would love to buy the farm, but we turned him down. ... Colonel Marston stops here. Linda stops here.

[Act] ask about himself \\\midrule

High Narcissism & Seastalker & [Ob] You are standing in a small room, that of the protective goddess Selkis. ... A beautiful ruby cluster, sparkling like fire, sits on the floor.

[Act] get ruby \\\midrule

Low Narcissism & Infidel & [Ob] You emerge into the warm night air of summer. ... A midget decked out in a Russian general's uniform is standing before the drinking fountain.

[Act] help midget  \\

\bottomrule
\end{tabular}
}
\caption{Qualitative examples of observation ([Ob]) \&  action ([Act]) pairs annotated with 16 personality types.}
\label{tab:personality_qual}
\scriptsize
\end{table*}

%% file: tables/compared_by_completion.tex
\begin{table*}[]
\scriptsize
\centering
\resizebox{0.8\linewidth}{!}{
\begin{tabular}{c|ccccc||c}

\toprule
\multicolumn{1}{c|}{\textbf{Game/Agent}} & \multicolumn{1}{c}{\textbf{NAIL}} & \multicolumn{1}{c}{\textbf{CALM}} & \multicolumn{1}{c}{\textbf{CMPS}} & \multicolumn{1}{l}{\textbf{CMPS+}} & \multicolumn{1}{c||}{\textbf{GALAD}} & \agentopeh \\ 
\cline{1-7}

\hline
BAL & 0.3 & 2.5 & 1.2 & 2.2 & 1.6 & 3.5 \\
BOR  & 1.4 & 3.6  & 3.3 & 3.7 & 3.5 & 3.3  \\
CUT  & 4.2  & 3.9 & 3.8  & 3.6  & 3.8  & 3.9 \\
DEA  & 0.8 & 1.6 & 1.6 & 1.7  & 1.8 & 1.3 \\
ENC  & 0.0 & 1.8 & 1.7 & 3.6 & 3.2 & 3.0 \\
HIT & 0.0 & 7.9 & 7.2 & 10.5 & 10.0 & 0.1 \\
HOL & 0.3 & 1.7 & 1.8 & 1.6 & 1.8 & 0.0 \\
INF & 0.1 & 0.4 & 0.4  & 0.4 & 0.4 & 0.1 \\
LUR & 0.0 & 0.4 & 0.8 & 0.3 & 0.3 & 0.0 \\
MOO & 7.1 & 9.3 & 9.3 & 8.2 & 10.9 & 7.6 \\
PLA & 0.5 & 1.6 & 1.3 & 1.6 & 2.2 & 1.9 \\
PLU & 1.0 & 2.7 & 2.8 & 2.8 & 3.2 & 5.5 \\
SEA & 1.0 & 3.4 & 4.4 & 3.9 & 4.4 & 7.3 \\
SOR & 0.5 & 2.6 & 2.6 & 2.6 & 1.8 & 5.2 \\
SPE  & 0.6  & 3.4 & 3.4 & 3.4 & 3.3 & 6.6 \\
STAR & -1.7 & -0.1 & -0.1 & -0.1 & 1.3 & 0.0  \\
STAT & 0.7 & 0.3 & 0.2 & 0.3 & 0.4 & 0.0 \\
SUS & 3.5 & 5.1 & 4.3 & 4.8 & 4.4 & 5.9 \\
SUSP & -1.7 & -0.7 & -0.8 & -0.4 & -0.7 & 0.0 \\
TRI & 0.1 & 1.6 & 1.6 & 1.5 & 1.6 & 6.9 \\
WIS & 0.3 & 5.0 & 5.1 & 5.0 & 5.2 & 6.0 \\
WIT & 2.8 & 9.2 & 8.6 & 9.2 & 9.9 & 11.4 \\
Z1 & -2.4 & 5.3 & 5.1 & 5.3 & 5.2 & 8.9 \\
Z2 & -2.5  & 2.5 & 4.0 & 2.5 & 2.4 & -0.2 \\
Z3 & 5.2 & 12.2 & 11.1 & 12.2 & 12.0 & 15.0 \\ 
\hline
\textbf{Average}  & 0.9 & 3.5 & 3.4 & 3.6 & 3.8 & 4.1 \\ 
\hline

\end{tabular}
}
\caption{Comparison with previous text-game adventure agents. We report \agentopeh\ as a representative example of the \oursys{} framework.}
\label{tab:baselines}
\end{table*}

%% file: tables/slam_detail_1.tex
\begingroup
\setlength{\tabcolsep}{1.0mm} %
\begin{table*}[h]
\centering
\small
\resizebox{\linewidth}{!}{
\begin{tabular}{c|c|cccccccccc}
\toprule
\textbf{Metric}& &\agentnp &\agentopeh &\agentopel &\agentconh &\agentconl &\agentexth &\agentextl &\agentagrh &\agentagrl \\
\midrule
\textit{Trajectory Length (\textdownarrow)} & - & 45.85$\pm$3.2 & 

57 $\pm$2.6 & 39.9 $\pm$2.1& 50.1 $\pm$5.2& 51.3 $\pm$8.2& 60.9 $\pm$20.1& 55.5 $\pm$3.5& 49.2 $\pm$3.3& 47.6$\pm$3.8\\
\cline{1-11}
\multirow{2}{*}{\rule[-0.1ex]{0pt}{5.1ex} \textit{Visit Count (\textuparrow)}} & \rule[-1ex]{0pt}{3.7ex} \textit{Com.} & 8.66$\pm$0.3 &

9.0 $\pm$0.4& 8.0 $\pm$0.1& 8.9 $\pm$0.3& 8.5 $\pm$0.3& 7.8 $\pm$0.9& 8.4 $\pm$0.4& 8.6 $\pm$0.3& 8.4$\pm$0.7\\
\rule[-1ex]{0pt}{3.5ex} &\textit{Unc.} &0.83$\pm$0.2 &1.20$\pm$0.4 &0.30$\pm$0.1 &0.89$\pm$0.1 &0.88$\pm$0.3 &0.67$\pm$0.4 &1.21$\pm$0.1 &0.82$\pm$0.2 &1.01$\pm$0.5 \\
\cline{2-11}

\cline{1-11}
\multirow{2}{*}{\rule[-0.1ex]{0pt}{4.9ex} \textit{Avg. Step (\textdownarrow)}} & \rule[-1ex]{0pt}{3.7ex} \textit{Com.} &12.64$\pm$2.1 &11.9 $\pm$1.0& 11.6 $\pm$0.4& 14.5 $\pm$0.6& 13.2 $\pm$0.9& 10.3 $\pm$1.7& 12.8 $\pm$0.2& 12.3 $\pm$0.6& 11.7$\pm$1.5\\
\rule[-0.6ex]{0pt}{3.1ex} &\textit{Unc.} & 8.62$\pm$3.5 & 6.4 $\pm$4.4& 12 $\pm$3.2& 17.5 $\pm$2.5& 13.5 $\pm$4.1& 6.1 $\pm$3.1& 16.7 $\pm$4.2& 9.4 $\pm$2.2& 12.3$\pm$6.0\\

\bottomrule

\end{tabular}
}
\caption{Full Results on Table~\ref{tab:slam} for \agentnp, \agentopeh, \agentopel , \agentconh, \agentconl, \agentexth, \agentextl, \agentagrh, and \agentagrl .}
\label{tab:slam_detail1}
\end{table*}
\endgroup

%% file: tables/slam_detail_2.tex
\begingroup
\setlength{\tabcolsep}{1.0mm} %
\begin{table*}[h]
\centering
\small
\resizebox{\linewidth}{!}{
\begin{tabular}{c|c|cccccccccc}
\toprule
\textbf{Metric}& &\agentnp &\agentneuh &\agentneul &\agentpsyh &\agentpsyl &\agentmach &\agentmacl &\agentnarh &\agentnarl \\
\midrule
\textit{Trajectory Length (\textdownarrow)} & - & 45.85$\pm$3.2 & 

48.9 $\pm$2.9& 54.7 $\pm$10.0& 50.4 $\pm$0.9& 44.9 $\pm$0.3& 46.1 $\pm$7.5& 53.7 $\pm$8.8& 48.7 $\pm$5.5& 46.2$\pm$3.6 \\
\cline{1-11}
\multirow{2}{*}{\rule[-0.1ex]{0pt}{5.1ex} \textit{Visit Count (\textuparrow)}} & \rule[-1ex]{0pt}{3.7ex} \textit{Com.} & 8.66$\pm$0.3 &

8.9 $\pm$0.2& 8 $\pm$0.7& 8.3 $\pm$0.6& 8.1 $\pm$0.2& 8.6 $\pm$0.6& 8.2 $\pm$0.2& 8.1 $\pm$0.5& 8.1$\pm$0.5\\
\rule[-1ex]{0pt}{3.5ex} &\textit{Unc.} &0.83$\pm$0.2 &1.2 $\pm$0.4& 0.6 $\pm$0.2& 0.8 $\pm$0.6& 0.3 $\pm$0.2& 1 $\pm$0.7& 1.1 $\pm$0.9& 0.6 $\pm$0.5& 0.5$\pm$0.1 \\
\cline{2-11}

\cline{1-11}
\multirow{2}{*}{\rule[-0.1ex]{0pt}{4.9ex} \textit{Avg. Step (\textdownarrow)}} & \rule[-1ex]{0pt}{3.7ex} \textit{Com.} &12.64$\pm$2.1 &13.8 $\pm$1.4& 10 $\pm$1.0& 13.3 $\pm$0.8& 12.3 $\pm$0.8& 12 $\pm$1.3& 13.7 $\pm$3.3& 12.4 $\pm$1.5& 12.1$\pm$0.5\\
\rule[-0.6ex]{0pt}{3.1ex} &\textit{Unc.} & 8.62$\pm$3.5 & 16.9 $\pm$1.8& 9 $\pm$5.7& 8.5 $\pm$4.0& 7.8 $\pm$2.9& 9.8 $\pm$4.7& 12.1 $\pm$2.6& 8.9 $\pm$6.5& 8.6$\pm$2.5\\

\bottomrule
\end{tabular}
}
\caption{Full Results on Table~\ref{tab:slam} for \agentnp, \agentneuh, \agentneul , \agentpsyh, \agentpsyl, \agentmach, \agentmacl, \agentnarh, and \agentnarl .}
\label{tab:slam_detail2}
\end{table*}
\endgroup

%% file: tables/gpt4_prompts.tex
\begin{table*}[!htp]
\centering
\small
\begin{tabular}{|p{0.9\textwidth}|}
\toprule
\textbf{GPT-4 annotation in \S~\ref{sec:walkthrough_analysis}.} \\
\midrule
VARIABLES: PERSONALITY, ACTION \\
\midrule
PROMPT\\
For given action, Determine whether the action exhibits high [PERSONALITY] or low [PERSONALITY] or is neutral with respect to [PERSONALITY].\\
You can choose from the following options, you should choose only one option, without any description. \\
Action: [ACTION]
\\
\\
\\

\toprule
\textbf{Acquiring 10 Personality Description in \S~\ref{sec:datamake}.} \\
\midrule
VARIABLES: PERSONALITY, DESCRIPTION \\
\midrule
PROMPT \\
Please paraphrase the following sentences describing the trait of [PERSONALITY]. \\ 
Generate 10 semantically distinct paraphrases:\\ 
- 5 paraphrases that emphasize high levels of the trait, and 5 paraphrases that emphasize low levels of the trait.\\ 
- Each paraphrase should reflect different aspects and nuances of the trait without overlapping.\\ 

Descriptions: [DESCRIPTION] 
\\
\\
\\

\toprule
\textbf{Acquiring 300 Diverse Situation in \S~\ref{sec:datamake}.} \\
\midrule
VARIABLES: -\\
\midrule
PROMPT \\ 
Generate 300 most common everyday places. \\
- Categorize them into 30 sub-categories, with 10 places in each category. \\
- List only the places without descriptions. \\
\\
\\
\\

\toprule
\textbf{Augmenting 5 detailed sentences in \S~\ref{sec:datamake}.} \\
\midrule
VARIABLES: PERSONALITY DESCRIPTION, PLACE \\
\midrule
PROMPT \\ 
Based on the following everyday place and personality description, generate 5 possible behaviors that this person might exhibit. \\
- Each behavior should be distinct and semantically different from the others. \\
- The behaviors should be plausible and realistic in the context of the given place and personality description. \\

Place: [PLACE]  \\
Personality Description: [PERSONALITY DESCRIPTION] \\

\\
\\

\bottomrule
\end{tabular}
\caption{Prompts that were used in our work.}
\label{tab:gpt4_prompts}

\end{table*}

%% file: tables/dataset_example.tex
\begin{table*}[ht]
\centering
\small
\begin{tabular}{l|p{6.2cm}|p{6.2cm}}
\toprule
\multicolumn{3}{c}{\textbf{Extraversion}} \\
\midrule
Valence & High & Low \\
\midrule
Place & Living Room & Living Room \\
\midrule
Description & I enjoy engaging in conversation with others frequently. & I often prefer to listen rather than speak in social settings. \\
\midrule
Data & 
\parbox[t]{6.2cm}{1. I start a lively discussion about the latest TV shows.\\2. I often host dinner parties to engage with different people.}
& 
\parbox[t]{6.2cm}{1. I quietly observe conversations rather than actively participating.\\2. I keep my comments brief when asked for my opinion.} \\
\bottomrule
\addlinespace[20pt]

\toprule
\multicolumn{3}{c}{\textbf{Openness}} \\
\midrule
Valence & High & Low \\
\midrule
Place & Living Room & Living Room \\
\midrule
Description & I often devise unique solutions and original creations. & I gravitate towards tasks that are consistent and unvarying. \\
\midrule
Data & 
\parbox[t]{6.2cm}{1. I create a multi-functional furniture piece for the living room.\\2. I design a unique piece of artwork for the living room wall.}
& 
\parbox[t]{6.2cm}{1. I always sit in the same spot on the couch.\\2. I stick to the same routine of cleaning the living room every Saturday morning.} \\
\bottomrule
\addlinespace[20pt]

\toprule
\multicolumn{3}{c}{\textbf{Conscientiousness}} \\
\midrule
Valence & High & Low \\
\midrule
Place & Living Room & Living Room \\
\midrule
Description & I work in a methodical and systematic manner. & I tend to overlook details and make mistakes due to a lack of attention. \\
\midrule
Data & 
\parbox[t]{6.2cm}{1. I follow a set cleaning routine for the living room each week.\\2. I systematically sort and arrange the DVD collection in alphabetical order.}
& 
\parbox[t]{6.2cm}{1. I often misplace the remote control due to not paying attention.\\2. I forget to water the plants regularly.} \\
\bottomrule
\addlinespace[20pt]

\toprule
\multicolumn{3}{c}{\textbf{Agreeableness}} \\
\midrule
Valence & High & Low \\
\midrule
Place & Living Room & Living Room \\
\midrule
Description & My interactions are marked by thoughtfulness and a gentle approach toward everyone. & I tend to be critical and often pinpoint others' shortcomings. \\
\midrule
Data & 
\parbox[t]{6.2cm}{1. I ask everyone's opinion before choosing a movie to watch.\\2. I speak softly not to disturb those who are reading or studying.}
& 
\parbox[t]{6.2cm}{1. I criticize the arrangement of the furniture.\\2. I point out the dust on the bookshelf.} \\
\bottomrule

\end{tabular}
\caption{Examples of intermediate results of the dataset creation process for Extraversion, Openness, Conscientiousness and Agreeableness.}
\label{tab:dataset_example1}
\end{table*}

%% file: tables/dataset_example2.tex
\begin{table*}[ht]
\centering
\small
\begin{tabular}{l|p{6.2cm}|p{6.2cm}}
\toprule
\multicolumn{3}{c}{\textbf{Neuroticism}} \\
\midrule
Valence & High & Low \\
\midrule
Place & Living Room & Living Room \\
\midrule
Description & I am prone to excessive worrying. & I generally maintain a relaxed demeanor, even under pressure. \\
\midrule
Data & 
\parbox[t]{6.2cm}{1. I fret about guests spilling drinks on the carpet.\\2. I worry about the kids damaging the furniture when they play.}
& 
\parbox[t]{6.2cm}{1. I calmly discuss disagreements without raising my voice.\\2. I comfortably entertain guests, not worrying about minor details.} \\
\bottomrule
\addlinespace[20pt]

\toprule
\multicolumn{3}{c}{\textbf{Machiavellianism}} \\
\midrule
Valence & High & Low \\
\midrule
Place & Living Room & Living Room \\
\midrule
Description & Sees others as pawns in their scheme, believing in the ease of manipulating most people. & Engages directly in conflicts instead of avoiding them for potential future gain. \\
\midrule
Data & 
\parbox[t]{6.2cm}{1. I convince others to move the furniture according to my preference.\\2. I manipulate others into agreeing with my TV program choices.}
& 
\parbox[t]{6.2cm}{1. I discuss the disagreement with my roommate openly instead of ignoring it.\\2. I speak up when I disagree with a friend's viewpoint.} \\
\bottomrule
\addlinespace[20pt]

\toprule
\multicolumn{3}{c}{\textbf{Narcissism}} \\
\midrule
Valence & High & Low \\
\midrule
Place & Living Room & Living Room \\
\midrule
Description & I hold a belief in my uniqueness, reinforced by frequent affirmations from others. & I tend to feel uncomfortable and uneasy when receiving praise or accolades from others. \\
\midrule
Data & 
\parbox[t]{6.2cm}{1. I decorate the living room to reflect my unique style.\\2. I always have the most unique and interesting stories to share.}
& 
\parbox[t]{6.2cm}{1. I deflect compliments by praising others.\\2. I downplay my achievements when they are brought up.} \\
\bottomrule
\addlinespace[20pt]

\toprule
\multicolumn{3}{c}{\textbf{Psychopathy}} \\
\midrule
Valence & High & Low \\
\midrule
Place & Living Room & Living Room \\
\midrule
Description & I believe retribution should be immediate and severe. & I do not seek to cause others to regret their actions towards me. \\
\midrule
Data & 
\parbox[t]{6.2cm}{1. I immediately remove a roommate's belongings from the living room if they upset me.\\2. I disconnect the WiFi as punishment if someone streams too much in the living room.}
& 
\parbox[t]{6.2cm}{1. I do not retaliate when my sibling uses my favorite chair without asking.\\2. I do not hold grudges when my friend spills drink on my carpet.} \\
\bottomrule
\end{tabular}
\caption{Examples of intermediate results of the dataset creation process for Neuroticism, Machiavellianism, Narcissism and Psychopathy.}
\label{tab:dataset_example2}
\end{table*}

%% file: tables/questionnaires_bfi.tex
\begin{table*}[!htp]
\centering
\small
\begin{tabular}{|p{0.9\textwidth}|}
\toprule
\textbf{Openness} \\
\midrule
I am original and come up with new ideas.  \\
I am curious about many different things.  \\
I am ingenious and a deep thinker.  \\
I have an active imagination.  \\
I am inventive.  \\
I value artistic and aesthetic experiences.  \\
I prefer work that is routine. (R) \\
I like to reflect and play with ideas.  \\
I have few artistic interests. (R) \\
I am sophisticated in art, music, or literature.  \\
\toprule
\textbf{Conscientiousness} \\
\midrule
I do a thorough job.   \\
I can be somewhat careless. (R)  \\
I am a reliable worker.   \\
I tend to be disorganized. (R)  \\
I tend to be lazy. (R)  \\
I persevere until the task is finished.   \\
I do things efficiently.   \\
I make plans and follow through with them.   \\
I am easily distracted. (R)  \\

\toprule
\textbf{Extraversion} \\
\midrule
I am talkative.   \\
I am reserved. (R)  \\
I am full of energy.   \\
I generate a lot of enthusiasm.   \\
I tend to be quiet. (R)  \\
I have an assertive personality.   \\
I am sometimes shy and inhibited. (R)  \\
I am outgoing and sociable.   \\

\toprule
\textbf{Agreeableness} \\
\midrule
I tend to find fault with others. (R)  \\
I am helpful and unselfish with others.   \\
I start quarrels with others. (R)  \\
I have a forgiving nature.   \\
I am generally trusting.   \\
I can be cold and aloof. (R)  \\
I am considerate and kind to almost everyone.   \\
I am sometimes rude to others. (R)  \\
I like to cooperate with others.   \\
\bottomrule
\end{tabular}
\caption{questionnaire items in BFI~\cite{John1999TheBF}. (R) indicates 'Reversed', which means a low tendency toward that personality trait.}
\label{tab:questionnaire_bfi}

\end{table*}

%% file: tables/questionnaire_sd3.tex
\begin{table*}[!htp]
\centering
\small
\begin{tabular}{|p{0.9\textwidth}|}

\toprule
\textbf{Psychopathy} \\
\midrule
I like to get revenge on authorities.   \\
I avoid dangerous situations. (R)  \\
Payback needs to be quick and nasty.   \\
People often say I’m out of control.   \\
It’s true that I can be mean to others.   \\
People who mess with me always regret it.   \\
I have never gotten into trouble with the law. (R)  \\
I enjoy having sex with people I hardly know.   \\
I’ll say anything to get what I want.   \\

\toprule
\textbf{Narcissism} \\
\midrule
People see me as a natural leader.   \\
I hate being the center of attention. (R)  \\
Many group activities tend to be dull without me.   \\
I know that I am special because everyone keeps telling me so.   \\
I like to get acquainted with important people.   \\
I feel embarrassed if someone compliments me. (R)  \\
I have been compared to famous people.   \\
I am an average person. (R)  \\
I insist on getting the respect I deserve.   \\
\toprule
\textbf{Machiavellianism} \\
\midrule
It’s not wise to tell your secrets.   \\
I like to use clever manipulation to get my way.   \\
Whatever it takes, you must get the important people on your side.   \\
Avoid direct conflict with others because they may be useful in the future.   \\
It’s wise to keep track of information that you can use against people later.   \\
You should wait for the right time to get back at people.   \\
There are things you should hide from other people to preserve your reputation.   \\
Make sure your plans benefit yourself, not others.   \\
Most people can be manipulated.  \\
\bottomrule
\end{tabular}
\caption{questionnaire items in SD-3~\cite{jones2014introducing}.}
\label{tab:questionnaire_sd3}

\end{table*}